\def\year{2022}\relax
\documentclass[letterpaper]{article} % DO NOT CHANGE THIS
\usepackage{aaai22_edited}  % DO NOT CHANGE THIS
\usepackage{times}  % DO NOT CHANGE THIS
\usepackage{helvet}  % DO NOT CHANGE THIS
\usepackage{courier}  % DO NOT CHANGE THIS
\usepackage[hyphens]{url}  % DO NOT CHANGE THIS
\usepackage{graphicx} % DO NOT CHANGE THIS
\usepackage{natbib}  % DO NOT CHANGE THIS AND DO NOT ADD ANY OPTIONS TO IT
\usepackage{caption} % DO NOT CHANGE THIS AND DO NOT ADD ANY OPTIONS TO IT
\DeclareCaptionStyle{ruled}{labelfont=normalfont,labelsep=colon,strut=off} % DO NOT CHANGE THIS
\usepackage{mixedgcImp}
\usepackage{algorithm}
\usepackage{algorithmic}

\nocopyright
\usepackage{newfloat}
\usepackage{listings}
\floatstyle{ruled}
\newfloat{listing}{tb}{lst}{}
\floatname{listing}{Listing}

\title{Online Missing Value Imputation and Change Point Detection \\with the Gaussian Copula}
\author{
     Yuxuan Zhao\thanks{Equal contribution},
      Eric Landgrebe\footnotemark[1], 
       Eliot Shekhtman, 
       Madeleine Udell \\
}
\affiliations{
    Cornell University, Ithaca, NY 14853, USA\\
    \{yz2295, ecl93,ess239,udell\}@cornell.edu
}

\usepackage{bibentry}
\begin{document}

\maketitle

\begin{abstract}
Missing value imputation is crucial for 
real-world data science workflows. 
Imputation is harder in the online setting, 
as it requires the imputation method itself to be able to evolve over time.
For practical applications, imputation algorithms should 
produce imputations that match the true data distribution, 
handle data of mixed types, including ordinal, boolean, and continuous variables, 
and scale to large datasets. 
In this work we develop a new online imputation algorithm for mixed data using the Gaussian copula. 
The online Gaussian copula model meets all the desiderata: 
its imputations match the data distribution even for mixed data, 
improve over its offline counterpart on the accuracy when the streaming data has a changing distribution,
and on the speed (up to an order of magnitude) especially on large scale datasets.
%and it scales well,
%achieving up to an order of magnitude speedup over its offline counterpart.
%The online algorithm can adapt to a changing data distribution in streaming data.
By fitting the copula model to online data,
we also provide a new method to detect change points in the multivariate dependence structure with missing values. 
Experimental results on synthetic and real world data validate the performance of the proposed methods.
\end{abstract}

\section{Introduction} 
Many modern datasets contain missing values;
yet many machine learning algorithms require complete data.
Hence missing value imputation is an important preprocessing step.
The progress in low rank matrix completion (LRMC) \citep{candes2010matrix, recht2010guaranteed} has 
led to widespread use in diverse applications \cite{bell2007lessons,yang2019oboe}.
LRMC succeeds when the data matrix can be well approximated by a low rank matrix.
While this assumption is often reasonable for sufficiently large data matrices \cite{udell2019big}, 
it usually fails when one dimension of the data matrix is much larger than the other.
We refer to such matrices as long skinny datasets,
or high rank matrices.% as in \citet{fan2017high}.
%Recent work has developed nonlinear imputation methods that improve on LRMC methods for long skinny datasets \cite{ongie2017algebraic, fan2019online, fan2020polynomial}.
When a long skinny dataset has \emph{mixed} type, 
consisting of a combination of 
ordinal, binary, and continuous (or real-valued) variables,
the imputation challenge is even greater,
and successful methods must account for the different distribution of each column.
% harder due to the dramatic scale difference among columns.
For example,
survey dataset may contain millions of respondents but only dozens of questions.
The questions may include both real-valued responses such as age and weight,
and ordinal responses on a Likert scale measuring how strongly a respondent agrees with certain stated opinions.
A Gaussian copula imputation model,
which adapts to the distribution of values in each column, has recently shown state-of-the-art performance on a variety of long skinny mixed datasets \cite{zhao2020missing}.
Our work builds on the success of this model, which we describe in greater detail below.

Missing values also appear in online data, generated by sensor networks, or ongoing surveys, 
as sensors fail or survey respondents fail to respond.
%which presents a substantial challenge for efficient data analysis,
%especially when missing value appears, 
In this setting, 
online (immediate) imputation for new data points is important to facilitate online decision-making processes.
%Despite rich literature on offline missing value imputation, online imputation methods are rather scarce.
However,
most missing value imputation methods, 
including missForest \cite{stekhoven2012missforest} and MICE \cite{buuren2010mice},
%and 
%GANs \cite{yoon2018gain, luo2018multivariate}, VAEs \cite{nazabal2020handling, fortuin2020gp} 
%deep generative imputation methods \cite{yoon2018gain, mattei2019miwae, nazabal2020handling},
%\cite{fortuin2020gp},
cannot easily update model parameters with 
new observation in the online setting.
Re-applying offline methods after seeing every new observation consumes too much time and space.
Online methods,
which incrementally update the model parameters every time new data is observed,
enjoy lower space and time costs and can adapt to changes in the data,
and hence are sometimes preferred even in the offline setting.

Another common interest for online data (or time series) is 
change point detection: does the data distribution change abruptly, 
and can we pinpoint when the change occurs?
While there are many different types of temporal changes, 
we focus on changes in the dependence structure of the data,
a crucial issue for many real world applications.
For example, 
classic Markowitz portfolio design uses
the dynamic correlation structure of exchange rates and market indexes 
to design a portfolio of assets that balances risk and reward \cite{markowitz1991foundations}.
%\cite{engle1990asset,cserban2007modelling}. 
In practice, the presence of missing values and mixed data handicaps most conventional change point detection approaches.

In this paper, we address all these challenges: 
our online algorithm can impute missing values and detect changes in the dependency structure 
of long skinny mixed data, including real-valued data and ordinal data as special cases.
Our online imputation method builds on the offline Gaussian copula imputation model  \cite{zhao2020missing}.
This model posits that each data point is generated by drawing a latent Gaussian vector.
%, with mean zero, unit variance, and correlation matrix $\Sigma$.
This latent Gaussian vector is then transformed to match the marginal distribution of each observed variable. 
%It is fully described by the marginal transformation function and the correlation matrix associated with the latent Gaussian vector.
%This is a natural model for mixed data, as 
Ordinals are assumed to result from 
thresholding a real-valued latent variable. In the case of, say, product ratings data, 
we can imagine the observed ordinal values result from thresholding the customer's (real-valued) affinity for a given product.
%Another major advantage of Gaussian copula imputation is that it uses no model hyperparameters.

%The learned correlation matrix $\Sigma$ can be used for imputation.% or to correlate variables. 

\paragraph{Contribution}
We make three major contributions:
(1) We propose an online algorithm for missing value imputation using the Gaussian copula model,
which 
incrementally updates the model and thus 
can adapt to a changing data distribution.
%It does not need to store historical data.
(2) %To fit the Gaussian copula more efficiently, even in the offline setting,
We develop a mini-batch Gaussian copula fitting algorithm to accelerate the training in the offline setting.
%and a parallel implementation.
Compared to the offline algorithm \cite{zhao2020missing},
our methods achieve nearly the same imputation accuracy but being an order of magnitude faster,
which allows the Gaussian copula model to scale to larger datasets.
(3) We propose a Monte Carlo test for dependence structure change detection \textit{at any time}.
%and show how to use the proposed test to detect change points with false discovery rate (FDR) control \textit{over all time}.
%that pinpoints changes in the correlation structure.
The method tracks the magnitude of the copula correlation update and 
reports a change point when the magnitude exceeds a threshold.
Inheriting the advantages of the Gaussian copula model, 
all our proposed methods naturally handle long skinny mixed data
with missing values, and have no model hyperparameters except for common online learning rate parameters.
This property is crucial in the online setting, where 
the best model hyperparameters may evolve.

\paragraph{Related work}
%\paragraph{Gaussian copula imputation}
The Gaussian copula has been used to impute incomplete mixed data in the offline setting 
%fits a Gaussian copula model from incomplete observations 
using an expectation maximization (EM) algorithm \cite{zhao2020missing}.
Here, we develop an online EM algorithm to incrementally update the copula correlation matrix,
following \cite{cappe2009line}, 
and an online method to estimates the marginals,
so that there is no need to store historical data except for the previous model estimate.
%Early work on our method was presented  in a non-archival workshop \cite{anonymous2020online}.

%\paragraph{Online missing value imputation}
Existing online imputation methods mostly rely on matrix factorization (MF).
Online LRMC methods \cite{balzano2010online, dhanjal2014online} assume a low rank data structure.
Consequently,
they work poorly for long skinny data. %as the low rank assumption generally fails \cite{udell2019big}.
Online KFMC \cite{fan2019online} first maps the data to a high dimensional space and assumes the mapped data has a low rank structure.
It learns a nonlinear structure and outperforms online LRMC for long skinny data.
However, its performance is sensitive to a selected rank $r$,
which should be several times larger than the data dimension $p$ and thus needs to be carefully tuned in a wide range.
As $p$ increases, it also requires increasing $r$ to outperform online LRMC methods; for moderate $p$, 
the $O(r^3)$ computation time of online KFMC becomes prohibitive.
For all aforementioned MF methods,
their underlying continuity assumptions can lead to poor performance on mixed data.
Moreover, 
the sensitivity to the rank poses a difficulty in the online setting,
as the best rank may vary over time, and the rank chosen 
by cross-validation early on can lead to poor performance or even divergence later.

While recent deep generative imputation methods \cite{yoon2018gain,mattei2019miwae} look like online methods (due to the SGD update), they actually require lots of data, and are slow to adapt to changes in the data stream, which are unsatisfying for real-time tasks.
Deep time series imputation methods \citep{cao2018brits,fortuin2020gp} use the future to impute the past, and thus do not suit the considered online imputation task.

%\paragraph{Online change point detection}
Change point detection (CPD) is an important topic with a long history. %on offline and online variants of the problem.
See \citet{aminikhanghahi2017survey} for an expansive review.
Online CPD seeks to identify change points in real-time, 
before seeing all the data.
Missing data is also a key challenge for CPD:
most CPD algorithms require complete data.
The simplest fix for this problem, 
imputation followed by a complete-data CPD method,
can hallucinate change points due to the changing missingness structure or imputation method used.
Our proposed method avoids these difficulties.
Another workaround, Bayesian online CPD methods \cite{adams2007bayesian, fearnhead2007line},
can fill out the missing entries by sampling from its posterior distribution given all observed entries.

\section{Methodology}
Gaussian copula has two parameters: the transformation function and the copula correlation matrix.
We first review Gaussian copula imputation with known model parameters.
Online imputation differs from offline imputation only in how we estimate the model parameters.
We assume the missing mechanism is missing completely at random (MCAR) throughout the paper, same as in the offline setting \cite{zhao2020missing},
but show our method is robust to missing not at random (MNAR) mechanism empirically in the supplement.
We show how to estimate the transformation online in \cref{sec:marginal_estimation} and how to estimate the copula correlation online in \cref{sec:online_correlation_estimation} with a given marginal estimate.% before presenting the online estimation of the copula correlation.
%the transformation online 
%Before presenting the online estimation of the copula correlation in \cref{sec:online_correlation_estimation}, 
%we 
%Different versions of implementation are discussed in \cref{sec:online_correlation_estimation}.

\paragraph{Notation}
Define $[p]=\{1,\ldots,p\}$ for $p\in \mathbb{N}^+$.
We use capital letters $\bX$ to denote matrices and lower-case letters $\bx$ to denote vectors. 
For a matrix $\bX$,
we refer to the $i$-th row, $j$-th column, and $(i,j)$-th entry as $\bx^i, \bX_j$ and $\xij$, respectively.
We use columns to represent variables and rows to represent examples. 
For a vector $\bx\in\Rbb^p$,
we use $\bx_{I}$ to denote the subvector of $\bx$ with entries in subset $I\subset [p]$.
For each row vector $\bx^i$, we use $\indexO_i, \indexM_i\subset [p]$ to denote the observed and missing locations respectively, 
and thus $\xio$ is observed and $\xim$ is missing.
We use $\phi(\cdot;\mu,\Sigma)$ for the PDF of a normal vector with mean $\mu$ and covariance matrix $\Sigma$.
\begin{figure*}[t]
\centering
\includegraphics[width=0.75\textwidth]{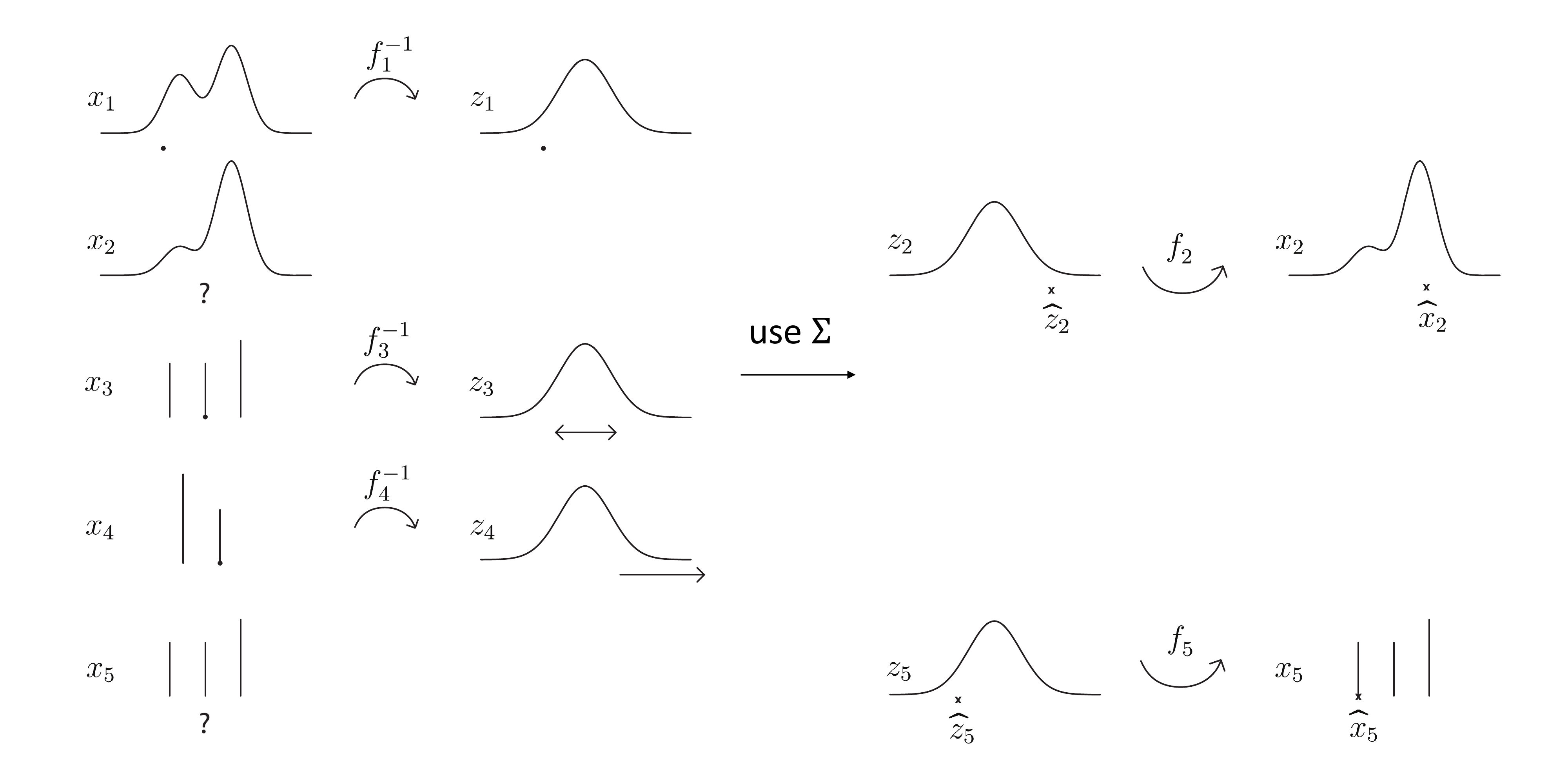} % Reduce the figure size so that it is slightly narrower than the column.
\caption{ Gaussian copula imputation for a 5-dim partially observed mixed vector.
Curves indicate the marginal probability density functions (for continuous) or probability mass function (for ordinal). 
First, compute the set of the latent normal vector which maps to the observation ($x_1,x_3$ and $x_4$) through $\bigf^{-1}$.
Second, compute the conditional mean of the latent normal vector at missing locations ($\hat z_2$ and $\hat z_5$) given the copula correlation $\Sigma$ and that $z_1, z_3$ and $z_4$ only take values from the computed inverse set.
Lastly, map the conditional mean through $\bigf$ to obtain the imputations $\hat x_2$ and $\hat x_5$.
}
\label{fig:imputation_example}
\end{figure*}
\paragraph{Gaussian copula imputation}
We now formally introduce the Gaussian copula model for mixed data \citep{hoff2007extending,fan2017high,feng2019high,zhao2020missing}.
We say a random vector $\bx \in \mathbb{R}^p$ follows the Gaussian copula model, $\bx \sim \gc$, if $\bx = \bigf(\bz):=(f_1(z_1),\ldots,f_p(z_p))$ with $\bz \sim N(\bo, \Sigma)$, 
for correlation matrix $\Sigma\in\Rbb^{p\times p}$ and elementwise monotone $\bigf: \Rbb^p \xrightarrow[]{} \Rbb^p$. % \cite{feng2019high,zhao2020missing}.
In other words, we generate a Gaussian copula random vector $\bx$ by first drawing a latent Gaussian vector $\bz$ with mean 0 and covariance $\Sigma$, and then applying the elementwise monotone function $\bigf$ to $\bz$ to produce $\bx$. 
If the cumulative distribution function (CDF) for $x_j$ is given by $F_j$, 
then $f_j$ is uniquely determined: $f_j = F_j^{-1} \circ \Phi$ where $\Phi$ is the standard Gaussian CDF. 
For ordinal $x_j$, 
the CDF $F_j$ and thus $f_j$ are step functions, so $f_j^{-1}(x_j):=\{z_j:f_j(z_j)=x_j\}$ is an interval.
If $\bx\sim \gc$ is observed at indices $\indexO$,
we map the conditional mean of $\zmis$ given observations $\xobs$ through $\bigf$ to impute the missing values $\xmis$ \cite{zhao2020missing}, 
as in \cref{eq:imputation} and visualized in \cref{fig:imputation_example}.
\begin{align}
    \hat \bx_{\indexM} &= \bigf_{\indexM}(\Ebb[\zmis|\xobs,\Sigma, \bigf])\nonumber\\
    &=\bigf_{\indexM}(\SigmaMO\SigmaOOinv\Ebb[\zobs|\xobs,\Sigma, \bigf]).
    \label{eq:imputation}
\end{align}
$\Sigma_{I,J}$ denotes the submatrix of $\Sigma$ with rows in $I$ and columns in $J$.
%We have
%$\Ebb[\zmis|\xobs,\Sigma, \bigf] =\SigmaMO\SigmaOOinv\Ebb[\zobs|\xobs,\Sigma, \bigf].$
The expectation $\Ebb[\zobs|\xobs,\Sigma, \bigf]$ is the mean of a normal vector $\zobs$ truncated to the region
$\prod_{j\in \indexO}f_j^{-1}(x_j)$,
which can be estimated efficiently.
%$\bigf^{-1}(\bx):=\prod_{j=1}^pf_j^{-1}(x_j)$,
%with $f_j^{-1}(x_j):=\Rbb$ if $x_j$ is missing.
%Such expectation can be estimated efficiently. 
For an incomplete matrix $\bX$, we assume that the rows are iid samples from $\gc$ with some $\bigf$ and $\Sigma$.
Each row is imputed separately as described above with $\bigf$ and $\Sigma$ replaced by their estimates. We now turn to the problem of estimating the parameters $\bigf$ and $\Sigma$.

\subsection{Online Marginal Estimation}
\label{sec:marginal_estimation}
In the offline setting, 
we estimate the transformation $\bigf$ based on the observed empirical distribution \cite{liu2009nonparanormal, zhao2020missing, zhao2020matrix}:
for $j\in [p]$, using observations in $\bX_j$, we construct the estimates as:
\begin{equation}
    \hat f_j =\hat{F}_j^{-1} \circ \Phi , \quad \hat f_j^{-1} = \Phi^{-1}\circ \hat{F}_j.
    \label{eq:marginal_estiamte}
\end{equation}
%where $\hat{F}_j$  and $\hat{F}_j^{-1}$ are the empirical CDF and empirical quantile function for observations in column $\bX_j$.
where $F_j$ and $F_j^{-1}$ are the empirical CDF and quantile function on the observed entries of the $j$-th variable.
In the online setting, we simply update the observation set as new data comes in for each column $\bX_j$.
Specifically,
we store a running window matrix $\tilde\bX\in\Rbb^{k\times p}$ which records the $k$ most recent observations for each column, 
and update  $\tilde\bX$ as new data comes in.
The window size is an online learning rate hyperparameter that should be tuned to improve accuracy. 
A longer window works better when the data distribution is mostly stable but has a few abrupt changes. 
If the data distribution changes rapidly, a shorter window is needed. Domain knowledge should also inform the choice of window length.
%Marginal estimates converge quickly: indeed, our experiments show a window size of $200$ generally suffices for accurate imputation.
%When the data dramatically vary over time, a smaller window size may be preferred.

\subsection{Online Copula Correlation Estimation}
\label{sec:online_correlation_estimation}
We estimate copula correlation matrix $\Sigma$ through maximum likelihood estimation (MLE).
The existing offline method \cite{zhao2020missing} applies EM algorithm to find the $\Sigma$ that maximizes the likelihood value. 
The key idea of our online estimation is to replace each offline EM iteration with an online EM variant, which incrementally updates the likelihood objective as new data comes in.
This online approach does not need to retain all data to perform updates.
We first present the offline likelihood objective to be maximized and then show how to update it in the online setting.

First when the data matrix $\bX$ is fully observed with all continuous columns,
we can compute exactly the Gaussian latent variable $\bz^i = \bigf^{-1}(\bx^i)\in \Rbb^p$ for each row $i$.
The likelihood objective is simply Gaussian likelihood $\sum_i\log\phi(\bz^i;0, \Sigma)$, 
and the MLE of the correlation matrix $\Sigma$ is the empirical correlation matrix $P_{\mathcal E}\left(\frac{1}{n} \sum_{i=1}^n  \bz^i(\bz^i)^\top\right)$,
where $P_{\mathcal E}$ scales its argument to output a correlation matrix:
     for $D = \text{diag}(\Sigma)$, $P_{\mathcal E}(\Sigma) = D^{-1/2} \Sigma D^{-1/2}$. 
%We now review a maximum likelihood estimation (MLE) approach for the copula correlation matrix $\Sigma$ in the offline setting  \cite{zhao2020missing}. 
%First, if the data matrix $\bX$ is fully observed with all continuous columns,we can compute exactly the Gaussian latent variable $\bz^i = \bigf^{-1}(\bx^i)\in \Rbb^p$ for each row $i$,
%and the problem reduces to the multivariate Gaussian setting with mean zero and unit variance.
%Thus the MLE of the correlation matrix $\Sigma$ is the empirical correlation matrix $P_{\mathcal E}\left(\frac{1}{n} \sum_{i=1}^n  \bz^i(\bz^i)^\top\right)$,
%where $P_{\mathcal E}$ scales its argument to output a correlation matrix:    for $D = \text{diag}(\Sigma)$, $P_{\mathcal E}(\Sigma) = D^{-1/2} \Sigma D^{-1/2}$. 
%When the data  $\bX$ is partially observed and has both continuous and ordinal columns, 

If there exist missing entries and ordinal variables in the data $\bX$,
the likelihood given observation $\xio$ is the integral over the latent Gaussian vector $\zio$ that maps to $\xio$ through $\bigf$: 
%under the marginal function $\bigf_{\indexO_i}$,
%i.e., $\zio \in \bigf^{-1}_{\indexO_i}(\xio)$ 
$z_j\in f_j^{-1}(x_j)$ for $j\in\indexO_i$
and $\bz^i_{\indexM_i}\in \Rbb^{|\indexM_i|}$.
For simplicity, we write $\bz^i\in \bigf^{-1}(\bx^i)$ by defining that $f_j^{-1}$ maps missing values to $\mathbb{R}$.
Thus the observed likelihood objective we seek to maximize (over $\Sigma$) is:
\begin{equation*}
    \ell(\Sigma; \{\xio\}_{i=1}^n) = \frac{1}{n} \sum_{i=1}^n \log \left( \int_{\bz^i \in \bigf^{-1}(\bx^i)} \phi(\bz^i;0,\Sigma)d\bz^i \right),
    \label{Eq: obs_likelihood}
\end{equation*}
The EM algorithm in \citet{zhao2020missing} avoids maximizing this difficult integral;
instead, the M-step at iteration $l$+1 maximizes the expectation of the complete likelihood $\ell(\Sigma;\{\bz^i,\xio\}_{i=1}^n)$, 
conditional on the observations $\xio$, the previous estimate $\Sigma^{l}$ and the marginal estimate $\hat \bigf$, computed in the E-step.
We denote this objective function as below:% $Q(\Sigma;\Sigma^l, \{\xio\}_{i=1}^n)$:
\begin{equation} 
    Q(\Sigma;\Sigma^l, \{\xio\}_{i=1}^n)=\frac{1}{n}\sum_{i=1}^n\mathbb{E}[\ell(\Sigma;\bz^i,\xio)| \bx^i_{\mathcal{O}_i}, \Sigma^l, \hat\bigf]\label{Eq: Q_function_}.
\end{equation}
Now the maximizer for \cref{Eq: Q_function_} is simply the expected ``empirical covariance matrix'' of the latent variables $\bz^i$: 
\begin{equation}
    \Sigma^{l+1} %= \argmax_{\Sigma} Q(\Sigma; \Sigma^t, X)
    = \sum_{i=1}^n \frac{1}{n}\mathbb{E}[\bz^i(\bz^i)^\top| \xio, \Sigma^l, \hat\bigf].
    \label{Eq:offline-Mstep}
\end{equation}
The expectation weights these $\bz^i$ by their conditional likelihood value.
%Using the fact that $\bz^i | \xio, \Sigma^l, \hat \bigf$ is a truncated normal vector,
These expectations are fast to approximate \cite{zhao2020missing, guo2015graphical}.
At last the the obtained estimate is scaled to have unit diagonal to satisfy the copula model constraints: $\Sigma^{l+1} \leftarrow P_{\mathcal E}\left(\Sigma^{l+1}\right)$.

Now we show how to adjust and maximize the objective $Q$ in the online setting. 
When data points come in different batches, 
i.e. rows $S_{t+1}$ observed at time $t+1$,
\citet{cappe2009line} propose to update the objective function $Q$ with new rows as:
\begin{equation}
    Q_{t+1}(\Sigma) = (1 - \gamma_{t})Q_{t}(\Sigma) + \gamma_{t}Q(\Sigma;\Sigma^t, \{\xio\}_{i\in S_{t+1}}),
    \label{eq:online_em}
\end{equation}
with $Q_1(\Sigma)=Q(\Sigma;\Sigma^0, \{\xio\}_{i\in S_1})$ given initial estimate $\Sigma^0$ and
a monotonically decreasing stepsize $\gamma_t \in (0,1)$.
Using \cref{eq:online_em}, we derive a very natural update rule,
stated as \cref{lemma:online_update}: 
in each step we simply take a weighted average of the previous covariance estimate and the estimate we get with a single EM step on the next batch of data.
We require the batch size to be larger than the data dimension $p$ to obtain a valid update.
One can still make an immediate prediction for each new data point, 
but to update the model we must wait to collect enough data 
or use overlapping data batches.
\begin{lemma} 		\label{lemma:online_update}
		For data batches $\{\bx^i\}_{i\in S_1},...,\{\bx^i\}_{i\in S_t}$ with $\bx^i\in \Rbb^p$ and $\min_{l\in[t]}|S_l|>p$, 
		and objective $Q_t(\Sigma)$ as in \cref{eq:online_em} for $\gamma_t\in (0,1)$.
		Given a marginal estimate $\hat \bigf$,
		for $l=1,\ldots,t$, $\Sigma^l\coloneqq \argmax_{\Sigma} Q_{l}(\Sigma)$ satisfies
\begin{equation}    \label{eq:online_update}
\Sigma^{t+1} =(1-\gamma_t)\Sigma^t + 
     \frac{\gamma_t}{|S_{t+1}|}\sum_{i \in S_{t+1}} \mathbb{E}[\bz^i(\bz^i)^\top| \bx^i_{\mathcal{O}_i}, \Sigma^t, \hat\bigf].
\end{equation}
\end{lemma}
We also project the resulting matrix to a correlation matrix as in the offline setting. 
The update takes $O(\alpha p^3 |S_t|)$ time with missing fraction $\alpha$ and $|S_t|$ rows.
The proof (in the supplement) shows that online EM formally requires 
a weighted update to the expectation computed in the E-step.
But for our problem, the parameter $\Sigma$, 
computed as the maximizer (in the M-step), 
is a \textit{linear} function 
of the computed expectation (from the E-step). 
Hence the maximizer also evolves according to the same simple 
weighted update.
A weighted update rule for the parameter fails --- leading to divergence --- for more general models, 
when the maximizer is not linear in the expectation,
such as for the low-rank-plus-diagonal copula correlation model of 
\citet{zhao2020matrix}.

\citet{cappe2009line} prove an online EM algorithm converges to the stationary points of the KL divergence between the true distribution of the observation $\pi$ (not necessarily the assumed model) and the learned model distribution,
under some regularity conditions.
We adapt their result to \cref{theorem:kl}.
% In our case, we obtain convergence under just two conditions on the distribution $\pi$;
% all other conditions hold automatically.
\begin{theorem}
Let $\pi(\xobs)$ be the distribution function of the true data-generating distribution of the observations and $g_{\Sigma}(\xobs)$ be the distribution function of the observed data from $\textup{GC}(\Sigma, \bigf)$, assuming data is missing uniformly at random (MCAR).
Suppose the step-sizes $\gamma_t\in (0,1)$ satisfy 
$\sum_{t=1}^\infty \gamma_t^2<\sum_{t=1}^\infty \gamma_t=\infty$.
Let $\mathcal{L}=\{\Sigma\in S_{++}^p: \nabla_{\Sigma}\textup{KL}(\pi||g_\Sigma)=0\}$ be the set of stationary points of $\textup{KL}(\pi||g_\Sigma)$ for a fixed $\bigf$.
%Denote by $\mathcal{L}=\{\Sigma\in S_{++}^p: \nabla_{\Sigma}\textup{KL}(\pi||g_\Sigma)=0\}$.
Under two regularity conditions on $\pi$ (see the supplement), 
%\[\lim_{t\xrightarrow[]{}\infty}\inf_{\Sigma \in \mathcal{L}}|\Sigma^{t} - \Sigma|=0 \mbox{ with probability }1, \quad \mbox{ for }\Sigma^t \mbox{ obtained in \cref{eq:online_update} with any fixed } \bigf.\]
the iterates $\Sigma^{t}$
produced by online EM (\cref{eq:online_update})
 converge to $\mathcal{L}$
 with probability $1$ as $t\xrightarrow{}\infty$.

% the distance from $\Sigma^{t}$, 
% obtained in \cref{eq:online_update}, to $\mathcal{L}$ converges to $0$ with probability $1$ as $t\xrightarrow{}\infty$.
\label{theorem:kl}
\end{theorem}
The conditions on stepsize $\gamma_t$ are standard for stochastic approximation methods.
If the true correlation $\Sigma$ generating the data evolves over time, 
a constant stepsize $\gamma_t\in(0,1)$ should be used to adapt the estimate to the changing correlation structure.
We find using $\gamma_i = c/(i+c)$ with $c=5$ for the offline setting and $\gamma_i = 0.5$ for the online setting gives good results throughout our experiments.
% Further tuning over different step size may bring additional gain.

\renewcommand{\algorithmicrequire}{\textbf{Input:}}
\renewcommand{\algorithmicensure}{\textbf{Output:}}
\begin{algorithm}[H]
\caption{Online Imputation with the Gaussian Copula}
\label{alg:onlie_imputation}
\begin{algorithmic}[1]
\REQUIRE Window size $k$, step size $\gamma_t$ for $t\in [T]$.%and time windows $S_t$ for 
\STATE Initialize $\Sigma^0$ and running window matrix $\tilde \bX\in\Rbb^{k\times p}$.
\FOR{$t=1,2,\ldots, T$}
\STATE Obtain new data batch $\{\bx^i\}_{i \in S_t}$, with $\bx^i$ partially observed at $\indexO_i$ and missing at $\indexM_i$. %\yxnote{Fix indention}
\STATE Replace the oldest point in $\tilde \bX_j$ with $x^i_j$ for $j\in \indexO_i, i\in S_t$.
\STATE Estimate marginals $\hat\bigf,\hat\bigf^{-1}$ using $\tilde \bX$ as in \cref{eq:marginal_estiamte} .
\STATE EM step update: 
obtain $\Sigma^{t+1}$ as in \cref{eq:online_update}.
\STATE Scale to a correlation matrix: $\Sigma^{t+1}= P_{\mathcal E}\left(\Sigma^{t+1}\right)$.
\STATE Impute $\hat\bx^i_{\indexM_i}$ using $\Sigma^{t+1}$ and $\hat \bigf$ as in \cref{eq:imputation} for $i\in S_t$. 
\ENDFOR
\ENSURE Imputation $\{\hat\bx^i_{\indexM_i}\}_{i\in S_t}$ and $\Sigma^t$ for $t\in[T]$. 
\end{algorithmic}
\end{algorithm}
\paragraph{Online versus offline implementation}
We may estimate $\hat \bigf$ in \cref{eq:online_update} either online or offline. The decision entails some tradeoffs.
When the storage limit is the main concern, as in the streaming data setting, 
we can employ the online marginal estimate, storing only a running window and a correlation matrix estimate.
We call such an implementation fully \textit{online EM}.
When the data marginal distribution evolves over time, 
it is also important to use \textit{online EM} to forget the old data.
On the other hand, when training time is the main concern but the whole dataset is available, 
%as in the offline data setting,
the online EM algorithm can be implemented as an offline mini-batch EM algorithm to accelerate convergence.
In that setting,
the offline marginals are used to provide more accurate and stable estimates as well as to reduce the time for estimating the marginals.
We call this implementation \textit{(offline) mini-batch EM}.
We present the fully online algorithm in \cref{alg:onlie_imputation} with 
data batches observed sequentially. 
%$\{\bx^i\}_{i \in S_1},\ldots, \{\bx^i\}_{i\in S_T}$ 
%consists of a single row at time $t$.
%The extension of \cref{alg:onlie_imputation} to the mini-batch setting ($|S_t|>1$) is straightforward.
%As in the offline setting, we can parallelize the online algorithm over all rows in $S_t$ to further accelerate it.
\paragraph{Parallelization}
Noting the computation of expectation in Eq. (\ref{Eq:offline-Mstep}) and \cref{eq:online_update} are separable over the rows, 
we have developed a parallel algorithm to accelerate the both the offline and the online EM algorithms.
For the long skinny datasets we target, 
this parallel algorithm allows for faster imputation
by exploiting multiple computational cores.

\subsection{Online Change Point Detection}
\label{sec:change point detection}
We first outline the change point detection (CPD) problem in the context of the Gaussian copula model.
Consider a sequence of incomplete mixed data observations  $\bx^1,\ldots,\bx^T \sim
\gc$, where $\bx^i$ is observed at locations $\indexO_i$ for $i\in[T]$. 
We wish to identify whether there is a change point $t_0$ ---
a time when the copula correlation  $\Sigma$ changes substantially --- 
and if so, when this change occurs.
%Moreover, in the online setting we wish to detect the change-point $t_0$ 
%as early as possible subject to a false alarm constraint. 
We formulate the single CPD problem as the following hypothesis test, for fixed $t_0$: $\bx^1,\ldots, \bx^{t_0} \sim \gc,  \mbox{ and }\bx^{t_0+1},\ldots, \bx^{T} \sim \textup{GC}(\tilde \Sigma, \bigf)$,
\begin{align}
\textup{H}_0: \tilde \Sigma = \Sigma \mbox{ versus } \textup{H}_1:\tilde \Sigma \neq \Sigma.
  \label{Eq:hypothesis}
\end{align}
We assume time-invariant marginal $\bigf$.
In practice, it suffices for $\bigf$ to be stable in a small local window.
The latent correlation matrix changes, reflecting the changing dependence structure.
To detect a change-point, a test statistic is  computed for each point to measure the deviation of new points from old distribution. 
A change is detected if the test statistic exceeds a certain threshold.
We consider the online detection problem instead of a retrospective analysis with all data available.
%although we will discuss how to extend our method to the retrospective setting.
Specifically, 
to test whether a change occurs at time $t_0$, 
we may use only the data $\bx^{t_0+1},\ldots,\bx^T$ for a small window length $T-t_0$ and the fitted model at time $t_0$.

To derive a test statistic,
notice that $\Sigma^{-1/2}\tilde\Sigma\Sigma^{-1/2}=\mathrm{I}_p$ under $H_0$.
Thus for some matrix norm $h$, we use the matrix distance 
$d(\Sigma, \tilde \Sigma; h) = h(\Sigma^{-1/2}\tilde\Sigma\Sigma^{-1/2}-\mathrm{I}_p)$ 
%between $\Sigma^{-1/2}\tilde\Sigma\Sigma^{-1/2}$ and $\mathrm{I}_p$ 
to measure the deviation of new points from old distribution.
While $\Sigma$ and $\tilde \Sigma$ are unknown, we replace them with the estimates $ \Sigma^{t_0}$ and $ \Sigma^T$, generated by the EM iteration up to time $t_0$ and time $T$, respectively.
Thus we construct our test statistic as $d(\Sigma^{t_0}, \Sigma^T;h)$: large values indicate high probability of a change point.
Experimentally, we find that different choices of $h$ give very similar trends.
Hence below we report results using the Frobenius norm as $h$, to reduce computation.

The change point is detected when $d(\Sigma^{t_0}, \Sigma^T;h)$ exceeds some threshold $b_\alpha$, 
which is chosen to control the false-alarm rate $\alpha$.
Calculating $b_\alpha$ analytically 
requires the asymptotic behaviour of the statistic under the null distribution,
which is generally intractable including our case.
We use Monte Carlo (MC) methods to simulate the null distribution of our test statistic and select the threshold. %using this simulated distribution.
This method is similar to the permutation test for CPD \cite{matteson2014nonparametric}.
We present our test for the hypothesis in \cref{Eq:hypothesis} as \cref{alg:change_point_detection} .
Notice comparing $d(\Sigma^{t_0}, \Sigma^T;h)$ to $b_\alpha$ is equivalent to comparing the returned empirical p-value with the desired false-alarm rate $\alpha$.
See  \cite{davison1997bootstrap, north2002note} for the use of empirical p-values.
In practice, 
$\alpha$ can be regarded as a hyperparameter to tune the false positive/negative rate. 
%For any fixed false-alarm rate $\alpha$,
%the decision on whether a change point occurs is deterministic by comparing the obtained p-values with $\alpha$,
%while the obtained p-value further provides how much trust we have on the decision.

\begin{algorithm}[h]
\caption{Monte Carlo test for Gaussian copula correlation change point detection}
\label{alg:change_point_detection}
\begin{algorithmic}[1]
\REQUIRE New data $\{\bx^i\}_{i=t_0+1}^T$,  the number of samples $B$, estimated model $\Sigma^{t_0}, \Sigma^T$ and $\bigf^{t_0}$.
\STATE Compute the test statistic $s=d(\Sigma^{t_0}, \Sigma^{T})$.
\FOR{$j=1,2,\ldots, B$}
\STATE Sample $\by^i\sim \textup{GC}(\Sigma^{t_0},\bigf^{t_0})$ and mask
$\by^{i}$ at where $\bx^{i+t_0}$ is missing for $i=1,...,T-t_0$.
%$\{\by^i\}_{i=1}^{T-t_0}$ where $\{\bx^i\}_{i=t_0+1}^T$ are missing.
\STATE Update the model at $t_0$ with new points $\{\by^i\}_{i=1}^{T-t_0}$.
\STATE Compute $s_j=d(\Sigma^{t_0}, \Sigma^{T,j})$ with the updated correlation $\Sigma^{T,j}$.
\ENDFOR
\ENSURE The p-value ${(|\{s_j:s\leq s_j\}|+1)}/{(B+1)}$.
\end{algorithmic}
\end{algorithm}

We have shown how to test if a change point happens at a time $t_0$. 
Repeating this test across time points may detect multiple change points, but also yield  many false positives. We discuss in the supplement how to alleviate this issue using recent development from online FDR \citep{javanmard2018online, ramdas2017online, ramdas2018saffron}.

\section{Experiments}
The experiments are divided into two parts: 
online datasets (rows obtained sequentially) 
and offline datasets (rows obtained simultaneously).
The online setting examine the ability of our methods to detect and learn the changing distribution of the steaming data. 
The  offline setting evaluate the speedups 
and the potential accuracy lost due to minibatch training 
and online marginal estimation compared to offline EM.
See the supplement for more experimental details and more experiments under different data dimension, missing ratio and missing mechanisms.
%All experiments use a laptop with a 3.1 GHz Intel Core i5 processor and 8 GB RAM. 
%See the supplement for more information including the algorithm implementation.
%The code for all the experiments and all used data are in the anonymous GitHub repository \url{https://anonymous.4open.science/r/89c86ff7-10c3-46e6-be9a-12c1b819ec74/}.

\textit{Algorithm implementation:}
we implement the \textit{offline EM} algorithm \cite{zhao2020missing},
the minibatch EM with online marginal estimate denoted by \textit{online EM}, and the minibatch EM with offline marginal estimate denoted by \textit{minibatch EM}.
For imputation comparison, we implement GROUSE \cite{balzano2010online} and online KFMC \cite{fan2019online}.
For fair comparison, we use 1 core for all methods, but report the acceleration brought by parallelism for all Gaussian copula methods in the supplement. 
We also implement the online Bayesian change point detection (BOCP) algorithm \cite{adams2007bayesian},
one of the best performing CPD method according to a recent evaluation paper \cite{van2020evaluation}.
The norm of subspace fitting residuals for GROUSE can also serve for CPD: a sudden peak of large residual norm indicates abrupt changes.
%Although it lacks a formal criterion to conduct CPD on the residuals, 
%it provides an intuitive visual illustration of the changes of the online model.
We compare our test statistic, defined in \cref{alg:change_point_detection}, with the residual norms from GROUSE, to see which identifies 
change points more accurately.

\textit{Tuning parameters selection:}
%All other methods incrementally update the model  using a mini batch at each iteration.
%Online imputation methods have two kinds of tuning parameters:
%model hyperparameters and the step size (or equivalently learning rate).
%We only tune over the  parameters that clearly influence the model's performance.
we do not use tuning parameter for offline EM and minibatch EM.
We use 1 tuning parameter for online EM:
the window size $m$ for online marginal estimates,
2 tuning parameters for GROUSE,
the rank and the step size,
and 2 tuning parameters for online KFMC,
the rank in a latent space and the regularization parameter.
%Grid search is used to choose all aforementioned tuning parameters.
%In practice, it is harder to find the best tuning parameter for GROUSE and online KFMC.
BOCP requires $4$ hyperparameters for its priors and its hazard function \cite{adams2007bayesian}.
%We choose them with a uniform random search \cite{bergstra2012random}.
%We employ training/validation/test split for all used datasets to conduct tuning parameter selection.
%See the supplement for more details and how we fix other tuning  parameters.

We note one other issue. For online algorithms, it is typical to choose hyperparameters during an initial ``burn-in'' period. 
For example, in GROUSE, choosing the step-size from initial data can result in divergence later on as the data distribution changes.
As a result, 
the maximum number of optimization iterations is also difficult to choose:
the authors' default settings are often insufficient to give good performance, while allowing too many iterations may lead to (worse) divergence.
We will report and discuss an example of divergence in our online real data experiment.

\textit{Imputation evaluation:}
for ordinal or real valued data,
we use mean absolute error (MAE) and root mean squared error (RMSE).
For mixed data,
we use the scaled MAE (SMAE), the MAE divided by the MAE of the median imputation.
A method that imputes better than the median has $\text{SMAE} <1$.

%We use this metric because it takes the imputation challenge for each column into account.
%For dataset whose columns have very different scales, the imputation performance for the column with the largest scale may dominate the overall performance. 

\paragraph{Offline synthetic  experiment}
\label{sec:experiment_synthetic_offline}
We construct a dataset consisting of $6000$ i.i.d. data points drawn from a 15-dimensional  Gaussian Copula, with $5$ continuous, %(exponential distribution with parameter $1/3$), 
$5$ ordinal with $5$ levels, and $5$ binary entries.
%$\bx^1,\ldots,\bx^n\sim \gc$ with $n=2000$,
%as in \cite{zhao2020missing}. 
We randomly mask $40\%$ entries as missing.
%The experiment is repeated over $10$ repetitions.
%For each repetition, we randomly sample the copula correlation matrix and the step points for generating ordinal columns which controls their probability mass functions,
%detailed in the Supplement.
%We evaluate the runtime, the average SMAE for each datatype, and the copula correlation estimation error for EM methods, computed as $||\hat \Sigma- \Sigma||_F/||\Sigma||_F$ for true correlation $\Sigma$ and estimate $\hat \Sigma$.
%We also evaluate the copula correlation estimation error, $||\hat \Sigma- \Sigma||_F/||\Sigma||_F$, for EM methods. %for true correlation $\Sigma$ and estimate $\hat \Sigma$.
Shown in \cref{table:offline_sim}, 
the minibatch and online variants of the EM algorithm converge 
substantially faster than offline EM
and provide similar imputation accuracy. 
The results are especially remarkable for online EM, 
which estimates the marginals using only $200$ points.
%With or without parallelism, 
The minibatch variant is three times faster than offline EM with the same accuracy. 
All EM methods outperform online KFMC and GROUSE, 
and even median imputation outperforms GROUSE.
Interestingly, the best rank for GROUSE is $1$.
The results here show LRMC methods fit poorly for long skinny datasets,
although the selected best rank, 1, 
misleadingly indicates the existence of low rank structure.

\begin{table}[htb]
\begin{center}
\begin{adjustbox}{width=\columnwidth}
%\begin{adjustbox}{width=\columnwidth}
\begin{tabular}{ |l | c |c |c |c |c|}
\hline
Method &  Runtime (s) & Continuous & Ordinal  & Binary 
 \\
\hline
Offline EM &  187.7(0.8) & 
0.79(.04) & 0.84(.03) & 0.63(.07) \\
Minibatch EM &  48.2(0.5) & 
0.79(.04) & 0.83(.03) & 0.63(.07) \\
Online EM &  54.5(3.4) & 
0.80(.04) & 0.84(.02) & 0.63(.07) \\
Online KFMC &  79.6(1.6) & 
0.92(.03)& 0.92(.02)& 0.67(.08)\\
GROUSE &  7.7(.3) &
1.17(.03) & 1.67(.05) & 1.10(.07) \\
\hline
\end{tabular}
\end{adjustbox}
\end{center}
\caption{Mean(sd) for runtime, imputation error of each data type for synthetic offline data over $10$ trials.}
\label{table:offline_sim}
%\end{adjustbox}
\end{table}

\paragraph{Online synthetic experiment}
Now we consider streaming data from a changing distribution.
%We evaluate the ability of online imputation algorithms to adapt to a changing correlation structure. 
To do this, we generate and mask the dataset similar to \cref{sec:experiment_synthetic_offline}, but set two change points at which a new correlation matrix is chosen:
$\bx^1,\ldots, 
\bx^t \sim \textup{GC}(\Sigma_1,\bigf)$, $\bx^{t+1},\ldots, \bx^{2t} \sim \textup{GC}(\Sigma_2,\bigf)$ and $\bx^{2t+1},\ldots, \bx^{3t} \sim \textup{GC}(\Sigma_3,\bigf)$, 
with $t=2000$.
%We repeat the experiment $10$ times.
%For each repetition, we randomly sample $\Sigma_1, \Sigma_2$ and $\Sigma_3$ and fix $\bigf$, as in the offline experiments.
We implement all online algorithms  from a cold start and 
make only one pass through the data, to mimic the streaming data setting. 
For comparison, we also implement offline EM and missForest \cite{stekhoven2012missforest} (also offline) and allow them to make multiple passes.

\begin{figure}
\centering
\includegraphics[width=\linewidth]{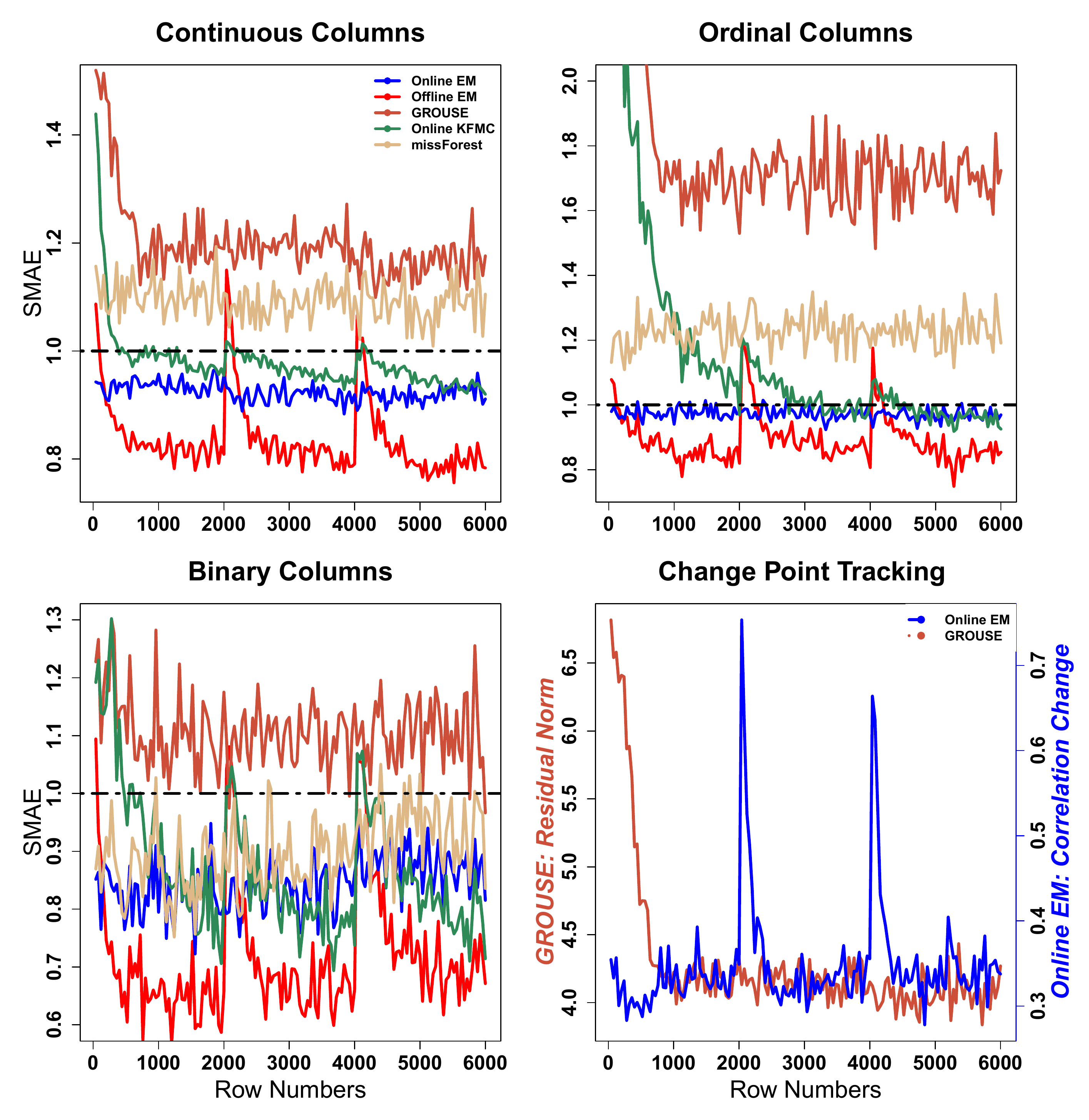} % Reduce the figure size so that it is slightly narrower than the column.
\caption{Mean imputation error and change point tracking statistics over $10$ trials for online synthetic datasets.
Each point stands for an evaluation over a data batch of $40$ points.}
\label{fig:online_synthetic}
\end{figure}

\begin{figure}[t]
\centering
\includegraphics[width=\linewidth]{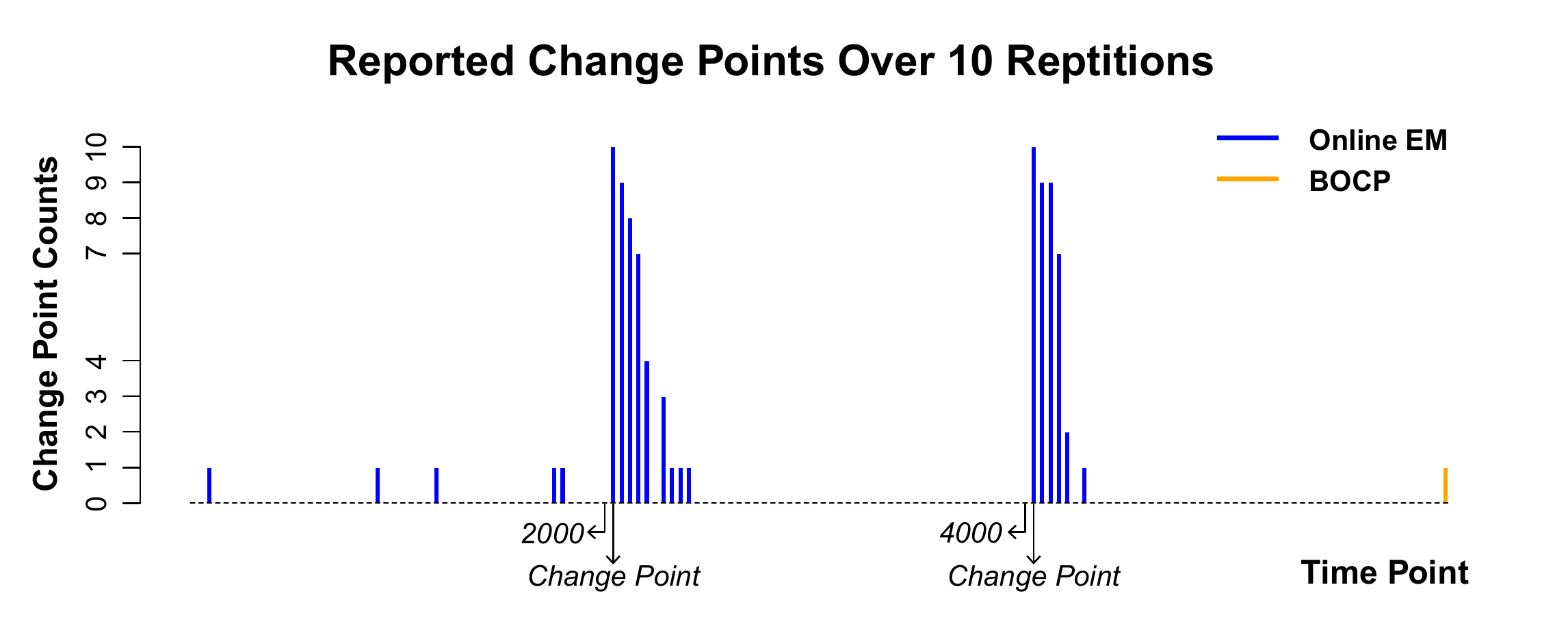} % Reduce the figure size so that it is slightly narrower than the column.
\caption{Change points from online EM detection (ours) and BOCP over $10$ trials in online synthetic experiments. Each bar stands for a decision over a data batch of $40$ points.}
\label{fig:sim_CPs}
\end{figure}

Shown in \cref{fig:online_synthetic},
online EM clearly outperforms the offline EM on average, 
by learning the changing correlation. 
Online EM has a sharp spike in error 
as the correlation abruptly shifts, 
but the error rapidly declines as it
learns the new correlation.
Both online EM and online KFMC outperform missForest.
Surprisingly, 
online KFMC cannot even outperform offline EM.
%which is only able to impute using a single correlation estimate for all data points.
GROUSE performs even worse in that it cannot outperform median imputation as in the offline setting.
%as a result it fails to outperform median imputation.
The results indicate online imputation methods can fail to learn the changing distribution when their underlying model does not fit the data well.
Our correlation deviation statistic provides accurate prediction for change points, while the residual norms from GROUSE remains stable after the burn-in period for model training, 
which verifies GROUSE cannot adapt to the changing dependence here.

Show in \cref{fig:sim_CPs}, online EM successfully detects both change points in all repetitions.
In fact, the algorithm detects a change point (of decreasing magnitude) during several batches after each true change, showing how long it takes to finally learn the new dependence structure.
To avoid the repeated false alarms, one could set a burn-in period following each detected change point.
In contrast, BOCP only reports one false discovery, showing its inability
to detect the changing dependence structure.

\paragraph{Offline real data experiment}
To further show the speedup of the minibatch algorithms,
we evaluate on a subset of the MovieLens 1M dataset \cite{harper2015movielens}
that consists of all movies with more than $1000$ ratings,
with $1$-$5$ ordinal ratings of size $6939\times 207$
with over $75\%$ entries missing. 
\cref{table:movielens}
shows that the minibatch and online EM still obtain comparable accuracy to the offline EM.
The minibatch EM is around 7 times faster than the offline EM. 
All EM methods significantly outperform online KFMC and GROUSE.
Interestingly, as the dataset gets wider, online KFMC loses its advantage over GROUSE.
The results here indicate the nonlinear structure learned by online KFMC fails to provide better imputation than the linear structure learned by GROUSE.
In contrast, the structural assumptions of our algorithm retain their advantage 
over GROUSE even on wider data.

%We report runtime and imputation error in \cref{table:movielens}.
\begin{table}
\begin{center}
\begin{tabular}{ |c|c| c|c| c| }
\hline
Method &  Runtime (s)& MAE & RMSE \\
\hline
Offline EM & 1690(9) & {0.583(.002)} &{0.883(.004)} \\ 
Minibatch EM & 252(2)& {0.585(.003)} & {0.886(.003)}\\
Online EM & 269(3)& 0.590(.002) & 0.890(.003) \\
Online KFMC& 176(21)& 0.631(.005) & 0.905(.006) \\
GROUSE &27(2) &0.634(.003) & 0.933(.004) \\
\hline
\end{tabular}
\end{center}
\label{table:movielens}
\caption{Mean(sd) for runtime and imputation error on a subset of MovieLens1M data over 10 trials.}
\end{table}

\paragraph{Online real data experiment}
\label{sec:experiments_online_realdata}
We now evaluate both imputation and CPD
on the daily prices and returns of 30 stocks currently in the Dow Jones Industrial Average (DJIA) across $5030$ trading days.
We consider two tasks: predicting each stock's price (or log return) today using only yesterday's data and a learned model.
After prediction, we reveal today's data to further update the model. %See the supplement for details.

% For each time $t$, the fitted model is used to provide out-of-sample imputation 
% for masked today's log returns.
% After today's log returns are revealed, 
% the model is updated using the complete $t$-th row.

\begin{figure}[t]
\centering
\includegraphics[width=\linewidth]{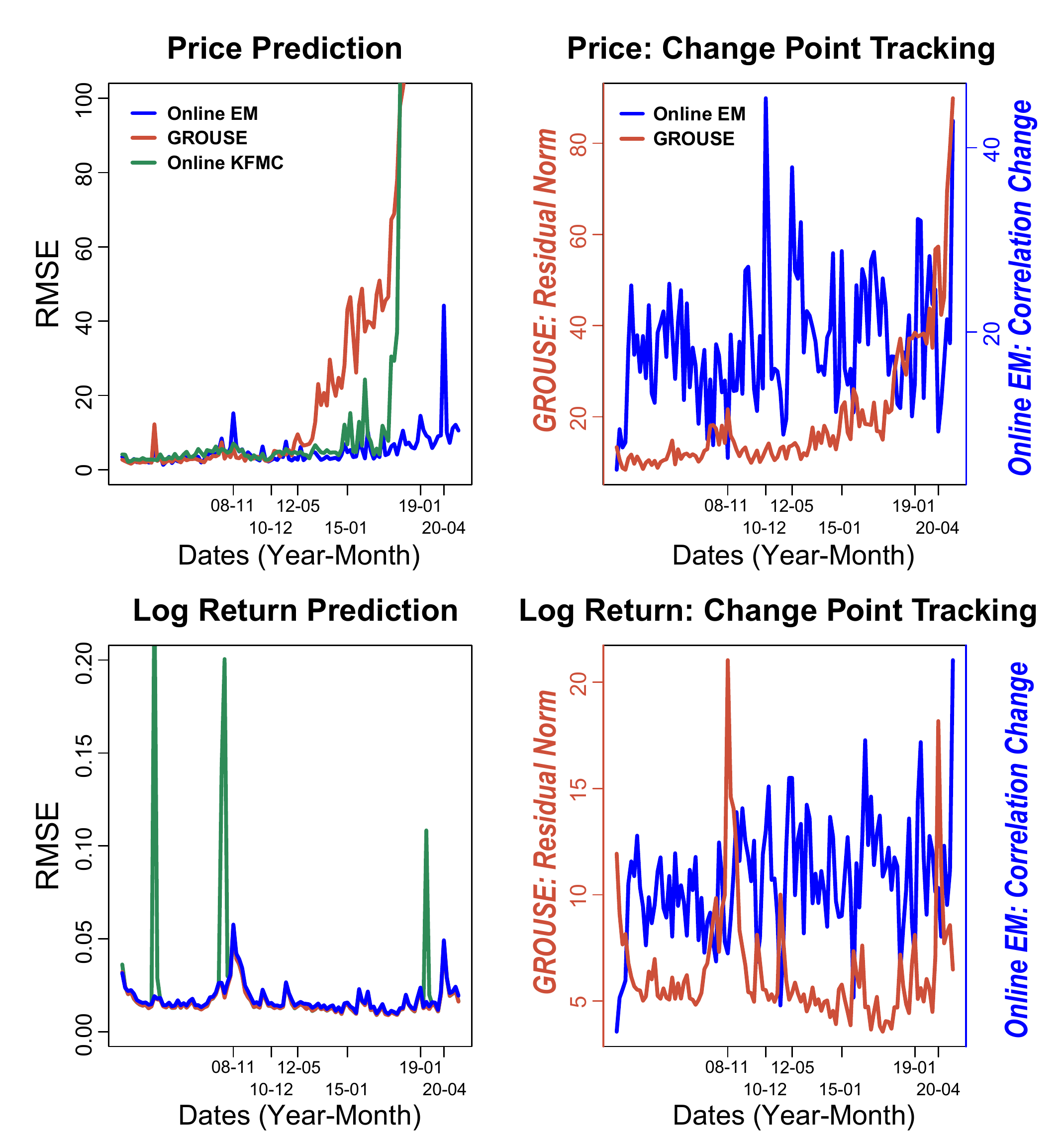} % Reduce the figure size so that it is slightly narrower than the column.
\caption{The imputation error and change tracking statistics for the DJIA daily price (left 2)  and the DJIA daily log-returns (right 2), averaged over all stocks.
Each point stands for an evaluation over a time interval of $40$ points.}
\label{fig:stocks}
\end{figure}

In \cref{fig:stocks},
the left 2 plots show that all methods predict prices well
early on, but
GROUSE and online KFMC both diverge eventually.
The residuals norm from GROUSE also indicate divergence.
%: GROUSE starting from around $2013$ and online KFMC starting from around $2017$.
In contrast, online EM has robust performance throughout.
Although the imputation error peaks around the start of $2020$, online EM is able to quickly adjust to the changing distribution: the imputation error quickly falls back.
Thus online EM stands out in that it obviates the need of online hyperparameter selection to have stable performance.
The right 2 plots show that online EM and GROUSE perform  similarly on log returns: their error curves almost overlap each other.
Online KFMC underperforms: it makes large errors more often.
We conjecture GROUSE and online KFMC perform better on the log returns than on 
the price data because the scale of the data is stable,
so that hyperparameters chosen early on still exhibit good performance later.
The good performance of GROUSE indicates the asset log returns are approximately low rank,
Still, online EM is robust to different (even changing) marginal data distributions and performs well on approximately low rank data.
%even when the data is approximately low rank.

As for CPD:
online EM shows similar results for both the price and log returns datasets, identifying fluctuating (often large) changes but no very distinct spikes on either dataset.
In fact, the algorithm classifies every (40-day) data batch as a 
change point, indicating the instability of stock data!
In contrast, GROUSE detects two large changes
in the log returns dataset and none in the price dataset. 
In the absence of ground-truth for change points, 
it is hard to compare the performance, but the improved stability 
of online EM to rescalings of the data is a clear advantage.
BOCP quickly diverged on both the price and the log-returns dataset and did not return meaning results before divergence.

\section{Conclusion}
We presented an online missing data imputation algorithm and change point detection method using  Gaussian copula for long skinny mixed datasets. 
The imputation performance can match or even exceed offline imputation,
and improves on  other state-of-the-art online imputations methods.
Our algorithm also provides speedup for offline Gaussian copula imputation.
The method can detect changes in the dependence structure, assuming the marginal remains stable over time.
%One important future direction would extend the online Gaussian copula estimation to wide datasets using low rank structure, %into the copula correlation matrix, 
%to ensure the computational complexity scales linearly in the number of observations,  generalizing the work of \citet{zhao2020matrix} for offline methods.
%A scalable offline method for fitting a low rank Gaussian copula is explored in \citet{zhao2020matrix}.
%Important future work remains: most notably, 
%key remaining problems include a method to identify changes in the marginal distribution,
Developing a method to identify changes in the marginal distribution is an important future work.
%and a less conservative online FDR control method 
%that supports online change point detection by utilizing the dependence structure of sequential tests.
%\mnote{Actually controlling FDR when p-values use MC simulation and hence are discrete.}
\clearpage

\section*{Acknowledgement}
The authors gratefully acknowledge support from NSF Award IIS-1943131, the ONR Young Investigator Program,
and the Alfred P. Sloan Foundation.
Special thanks to Xiaoyi Zhu for her assistance in creating our figures.
\bibliography{aaai22}

\begin{thebibliography}{37}
\providecommand{\natexlab}[1]{#1}

\bibitem[{Adams and MacKay(2007)}]{adams2007bayesian}
Adams, R.~P.; and MacKay, D.~J. 2007.
\newblock Bayesian online changepoint detection.
\newblock \emph{arXiv preprint arXiv:0710.3742}.

\bibitem[{Aminikhanghahi and Cook(2017)}]{aminikhanghahi2017survey}
Aminikhanghahi, S.; and Cook, D.~J. 2017.
\newblock A survey of methods for time series change point detection.
\newblock \emph{Knowledge and information systems}, 51(2): 339--367.

\bibitem[{Balzano, Nowak, and Recht(2010)}]{balzano2010online}
Balzano, L.; Nowak, R.; and Recht, B. 2010.
\newblock Online identification and tracking of subspaces from highly
  incomplete information.
\newblock In \emph{2010 48th Annual allerton conference on communication,
  control, and computing (Allerton)}, 704--711. IEEE.

\bibitem[{Bell and Koren(2007)}]{bell2007lessons}
Bell, R.~M.; and Koren, Y. 2007.
\newblock Lessons from the netflix prize challenge.
\newblock \emph{Acm Sigkdd Explorations Newsletter}, 9(2): 75--79.

\bibitem[{Bergstra and Bengio(2012)}]{bergstra2012random}
Bergstra, J.; and Bengio, Y. 2012.
\newblock Random search for hyper-parameter optimization.
\newblock \emph{Journal of machine learning research}, 13(2).

\bibitem[{Buuren and Groothuis-Oudshoorn(2010)}]{buuren2010mice}
Buuren, S.~v.; and Groothuis-Oudshoorn, K. 2010.
\newblock mice: Multivariate imputation by chained equations in R.
\newblock \emph{Journal of statistical software}, 1--68.

\bibitem[{Candes and Plan(2010)}]{candes2010matrix}
Candes, E.~J.; and Plan, Y. 2010.
\newblock Matrix completion with noise.
\newblock \emph{Proceedings of the IEEE}, 98(6): 925--936.

\bibitem[{Cao et~al.(2018)Cao, Wang, Li, Zhou, Li, and Li}]{cao2018brits}
Cao, W.; Wang, D.; Li, J.; Zhou, H.; Li, Y.; and Li, L. 2018.
\newblock BRITS: bidirectional recurrent imputation for time series.
\newblock In \emph{Proceedings of the 32nd International Conference on Neural
  Information Processing Systems}, 6776--6786.

\bibitem[{Capp{\'e} and Moulines(2009)}]{cappe2009line}
Capp{\'e}, O.; and Moulines, E. 2009.
\newblock On-line expectation--maximization algorithm for latent data models.
\newblock \emph{Journal of the Royal Statistical Society: Series B (Statistical
  Methodology)}, 71(3): 593--613.

\bibitem[{Davison and Hinkley(1997)}]{davison1997bootstrap}
Davison, A.~C.; and Hinkley, D.~V. 1997.
\newblock \emph{Bootstrap methods and their application}.
\newblock 1. Cambridge university press.

\bibitem[{Dhanjal, Gaudel, and Cl{\'e}men{\c{c}}on(2014)}]{dhanjal2014online}
Dhanjal, C.; Gaudel, R.; and Cl{\'e}men{\c{c}}on, S. 2014.
\newblock Online matrix completion through nuclear norm regularisation.
\newblock In \emph{Proceedings of the 2014 SIAM International Conference on
  Data Mining}, 623--631. SIAM.

\bibitem[{Fan et~al.(2017)Fan, Liu, Ning, and Zou}]{fan2017high}
Fan, J.; Liu, H.; Ning, Y.; and Zou, H. 2017.
\newblock High dimensional semiparametric latent graphical model for mixed
  data.
\newblock \emph{Journal of the Royal Statistical Society. Series B: Statistical
  Methodology}, 79(2): 405--421.

\bibitem[{Fan and Udell(2019)}]{fan2019online}
Fan, J.; and Udell, M. 2019.
\newblock Online high rank matrix completion.
\newblock In \emph{Proceedings of the IEEE Conference on Computer Vision and
  Pattern Recognition}, 8690--8698.

\bibitem[{Fearnhead and Liu(2007)}]{fearnhead2007line}
Fearnhead, P.; and Liu, Z. 2007.
\newblock On-line inference for multiple changepoint problems.
\newblock \emph{Journal of the Royal Statistical Society: Series B (Statistical
  Methodology)}, 69(4): 589--605.

\bibitem[{Feng and Ning(2019)}]{feng2019high}
Feng, H.; and Ning, Y. 2019.
\newblock High-dimensional mixed graphical model with ordinal data: Parameter
  estimation and statistical inference.
\newblock In \emph{The 22nd International Conference on Artificial Intelligence
  and Statistics}, 654--663.

\bibitem[{Fortuin et~al.(2020)Fortuin, Baranchuk, R{\"a}tsch, and
  Mandt}]{fortuin2020gp}
Fortuin, V.; Baranchuk, D.; R{\"a}tsch, G.; and Mandt, S. 2020.
\newblock Gp-vae: Deep probabilistic time series imputation.
\newblock In \emph{International Conference on Artificial Intelligence and
  Statistics}, 1651--1661. PMLR.

\bibitem[{Guo et~al.(2015)Guo, Levina, Michailidis, and Zhu}]{guo2015graphical}
Guo, J.; Levina, E.; Michailidis, G.; and Zhu, J. 2015.
\newblock Graphical models for ordinal data.
\newblock \emph{Journal of Computational and Graphical Statistics}, 24(1):
  183--204.

\bibitem[{Harper and Konstan(2015)}]{harper2015movielens}
Harper, F.~M.; and Konstan, J.~A. 2015.
\newblock The movielens datasets: History and context.
\newblock \emph{Acm transactions on interactive intelligent systems (tiis)},
  5(4): 1--19.

\bibitem[{Hoff et~al.(2007)}]{hoff2007extending}
Hoff, P.~D.; et~al. 2007.
\newblock Extending the rank likelihood for semiparametric copula estimation.
\newblock \emph{The Annals of Applied Statistics}, 1(1): 265--283.

\bibitem[{Javanmard, Montanari et~al.(2018)}]{javanmard2018online}
Javanmard, A.; Montanari, A.; et~al. 2018.
\newblock Online rules for control of false discovery rate and false discovery
  exceedance.
\newblock \emph{The Annals of statistics}, 46(2): 526--554.

\bibitem[{Liu, Lafferty, and Wasserman(2009)}]{liu2009nonparanormal}
Liu, H.; Lafferty, J.; and Wasserman, L. 2009.
\newblock The nonparanormal: Semiparametric estimation of high dimensional
  undirected graphs.
\newblock \emph{Journal of Machine Learning Research}, 10(10).

\bibitem[{Markowitz(1991)}]{markowitz1991foundations}
Markowitz, H.~M. 1991.
\newblock Foundations of portfolio theory.
\newblock \emph{The journal of finance}, 46(2): 469--477.

\bibitem[{Mattei and Frellsen(2019)}]{mattei2019miwae}
Mattei, P.-A.; and Frellsen, J. 2019.
\newblock MIWAE: Deep generative modelling and imputation of incomplete data
  sets.
\newblock In \emph{International Conference on Machine Learning}, 4413--4423.
  PMLR.

\bibitem[{Matteson and James(2014)}]{matteson2014nonparametric}
Matteson, D.~S.; and James, N.~A. 2014.
\newblock A nonparametric approach for multiple change point analysis of
  multivariate data.
\newblock \emph{Journal of the American Statistical Association}, 109(505):
  334--345.

\bibitem[{North, Curtis, and Sham(2002)}]{north2002note}
North, B.~V.; Curtis, D.; and Sham, P.~C. 2002.
\newblock A note on the calculation of empirical P values from Monte Carlo
  procedures.
\newblock \emph{The American Journal of Human Genetics}, 71(2): 439--441.

\bibitem[{Pagotto(2019)}]{ocp}
Pagotto, A. 2019.
\newblock \emph{ocp: Bayesian Online Changepoint Detection}.
\newblock R package version 0.1.1.

\bibitem[{Ramdas et~al.(2017)Ramdas, Yang, Wainwright, and
  Jordan}]{ramdas2017online}
Ramdas, A.; Yang, F.; Wainwright, M.~J.; and Jordan, M.~I. 2017.
\newblock Online control of the false discovery rate with decaying memory.
\newblock In \emph{Proceedings of the 31st International Conference on Neural
  Information Processing Systems}, 5655--5664.

\bibitem[{Ramdas et~al.(2018)Ramdas, Zrnic, Wainwright, and
  Jordan}]{ramdas2018saffron}
Ramdas, A.; Zrnic, T.; Wainwright, M.; and Jordan, M. 2018.
\newblock SAFFRON: an adaptive algorithm for online control of the false
  discovery rate.
\newblock In \emph{International conference on machine learning}, 4286--4294.
  PMLR.

\bibitem[{Recht, Fazel, and Parrilo(2010)}]{recht2010guaranteed}
Recht, B.; Fazel, M.; and Parrilo, P.~A. 2010.
\newblock Guaranteed minimum-rank solutions of linear matrix equations via
  nuclear norm minimization.
\newblock \emph{SIAM review}, 52(3): 471--501.

\bibitem[{Robertson et~al.(2019)Robertson, Wildenhain, Javanmard, and
  Karp}]{robertson2019onlinefdr}
Robertson, D.~S.; Wildenhain, J.; Javanmard, A.; and Karp, N.~A. 2019.
\newblock onlineFDR: an R package to control the false discovery rate for
  growing data repositories.
\newblock \emph{Bioinformatics}, 35(20): 4196--4199.

\bibitem[{Stekhoven and B{\"u}hlmann(2012)}]{stekhoven2012missforest}
Stekhoven, D.~J.; and B{\"u}hlmann, P. 2012.
\newblock MissForest—non-parametric missing value imputation for mixed-type
  data.
\newblock \emph{Bioinformatics}, 28(1): 112--118.

\bibitem[{Udell and Townsend(2019)}]{udell2019big}
Udell, M.; and Townsend, A. 2019.
\newblock Why are big data matrices approximately low rank?
\newblock \emph{SIAM Journal on Mathematics of Data Science}, 1(1): 144--160.

\bibitem[{van~den Burg and Williams(2020)}]{van2020evaluation}
van~den Burg, G.~J.; and Williams, C.~K. 2020.
\newblock An evaluation of change point detection algorithms.
\newblock \emph{arXiv preprint arXiv:2003.06222}.

\bibitem[{Yang et~al.(2019)Yang, Akimoto, Kim, and Udell}]{yang2019oboe}
Yang, C.; Akimoto, Y.; Kim, D.~W.; and Udell, M. 2019.
\newblock OBOE: Collaborative filtering for AutoML model selection.
\newblock In \emph{Proceedings of the 25th ACM SIGKDD International Conference
  on Knowledge Discovery \& Data Mining}, 1173--1183.

\bibitem[{Yoon, Jordon, and Schaar(2018)}]{yoon2018gain}
Yoon, J.; Jordon, J.; and Schaar, M. 2018.
\newblock Gain: Missing data imputation using generative adversarial nets.
\newblock In \emph{International Conference on Machine Learning}, 5689--5698.
  PMLR.

\bibitem[{Zhao and Udell(2020{\natexlab{a}})}]{zhao2020matrix}
Zhao, Y.; and Udell, M. 2020{\natexlab{a}}.
\newblock Matrix Completion with Quantified Uncertainty through Low Rank
  Gaussian Copula.
\newblock \emph{Advances in Neural Information Processing Systems}, 33.

\bibitem[{Zhao and Udell(2020{\natexlab{b}})}]{zhao2020missing}
Zhao, Y.; and Udell, M. 2020{\natexlab{b}}.
\newblock Missing Value Imputation for Mixed Data via Gaussian Copula.
\newblock In \emph{Proceedings of the 26th ACM SIGKDD International Conference
  on Knowledge Discovery \& Data Mining}, 636--646.

\end{thebibliography}
 \begin{appendix}
\clearpage

\section{Online change point detection algorithm}
In the main paper we introduced a method to determine whether a change point occurs at a time $t_0$ using the new data $\{\bx^i\}_{i=t_0+1}^T$.
% Notice that our statistic only uses the estimated correlation matrix at the beginning $t_0$ and end $T$ of 
% the batch; so in fact, it also tests the hypothesis 
% that a chance point occurred at any time between $t_0$ and $T$. 
In the online setting, we will seek to detect whether 
a change point occurred at the start of any of the time intervals
$S_1, \ldots, S_T$ using the corresponding data batches $\{\bx^i\}_{i \in S_1},\ldots, \{\bx^i\}_{i\in S_T}$.
(Notice that we can detect a change point at any time by using 
overlapping time windows.)
To do so, we can apply our MC test to every new data batch $\{\bx^i\}_{i \in S_t}$ for $t\in[T]$.
Unfortunately, controlling the significance level for each test still yields too many false positives due to the number of tests $T$.
Instead, we use online FDR control methods \cite{ramdas2017online,javanmard2018online, ramdas2018saffron} to control the FDR to a specified level across the whole process.
At each time point, the decision as to whether a change point occurs is made by comparing the obtained p-values from the MC test with the allocated time-specific significance level, 
which only depends on previous decisions.
We summarize the sequential algorithm in \cref{alg:online_change_point}.

\begin{algorithm}[h]
\caption{Online change point detection via Gaussian copula}
\label{alg:online_change_point}
\begin{algorithmic}[1]
    \REQUIRE Online FDR control algorithm $\mathcal{A}$, 
time windows $S_1,\ldots, S_T$, 
the FDR level $\alpha$, 
and an initial model estimate $\Sigma^0$ and $\bigf^0$.
\FOR{$t=1,2,\ldots, T$}
\STATE Obtain new data batch $\{\bx^i\}_{i \in S_t}$ 
\STATE Update the model estimate as $\Sigma^{t}$ and $\bigf^t$ as in our Algorithm 1 (the main paper).
\STATE Obtain the Monte carlo p-value $p_t$  with new data $\{\bx^i\}_{i \in S_t}$, $\Sigma^{t-1},\Sigma^t$ and $\bigf^{t-1}$ as input.
\STATE Obtain the significance  $\alpha_t=\mathcal{A}(\{R_1,\ldots, R_{t-1}\})$.
\STATE Set the binary decision $R_t=1$ if $p_t<\alpha_t$ and $0$ otherwise.
\ENDFOR
\ENSURE Decisions $\{R_t\}_{t\in[T]}$ and p-values $\{p_t\}_{t\in[T]}$.
\end{algorithmic}
\end{algorithm}

If the time intervals are very short,
it can be useful to set a burn-in period to estimate the new correlation matrix 
before attempting to detect more change points in order to prevent false positives.
Our method as stated is designed to detect change points quickly by 
using the model at the end of the time interval 
as an estimate for the new model $\tilde \Sigma$. 
Instead, one could wait longer to compute a better estimate of $\tilde \Sigma$
to assess whether a change point had occurred in an earlier interval. 
This approach detects change points slower and requires more memory than our approach, but could deliver higher precision.

We point out an important practical concern:
online FDR control methods often allocate very small significance levels ($<10^{-4}$) in practice \cite{robertson2019onlinefdr}, while the smallest p-value that the MC test with $B$ samples can output is $1/(B+1)$.
Under the null that no change point happens, 
the probability that the test statistics $s$ is larger than all $\{s_j\}_{j\in [B]}$ computed from MC samples is $1/(B+1)$.
Thus only when $B$ is very large ($>10^4$)
can a MC test possibly detect a change point.
Setting $B$ this large is usually computationally prohibitive.
One ad-hoc remedy is to use the biased empirical p-value $\#\{:s\leq s_j\}/(B+1)$,
which can output a p-value of 0;
however, this approach is equivalent to choose all significance levels $\alpha_t\in  (\frac{1}{B+1}, \frac{2}{B+1})$.
Developing a principled approach to online FDR control 
that can target less conservative significance levels (higher power) in the context of online CPD constitutes important future work.

\section{Experimental details}
All our used algorithms and experiments are in an anonymous Github repo \footnote{\url{https://anonymous.4open.science/r/Online-Missing-Value-Imputation-and-Change-Point-Detection-with-the-Gaussian-Copula-7308/}}.
\paragraph{Algorithm implementaion}
All our implementation codes are provided in the supplementary materials.
All experiments use a laptop with a 3.1 GHz Intel Core i5 processor and 8 GB RAM.
Our EM methods are implemented using Python. 
We implement GROUSE and online KFMC using the authors' provided Matlab codes at \url{https://web.eecs.umich.edu/~girasole/grouse/} and \url{https://github.com/jicongfan/Online-high-rank-matrix-completion}.
Bayesian online change point (BOCP) is implemented using the R package \texttt{ocp} \cite{ocp}.

\paragraph{Tuning parameter selection}
For all methods but BOCP, we use grid search to choose the tuning parameters.
For BOCP, we use uniform random search to choose the tuning parameters \cite{bergstra2012random}.

The window size of online EM is selected from $\{50,100,200\}$ for online experiments and fixed as $200$ for offline experiments.
The constant $c$ in step size $c/t$ is selected from $\{0.1, 1, 10\}$ for GROUSE on offline experiments. The constant step size $c$ is selected from $\{10^{-8}, 10^{-4}, 10^{-2}\}$ for GROUSE on online experiments.
The rank is selected from $\{1,5,10\}$ for GROUSE on all experiments.
The rank is selected from $\{200,300,400\}$ and the regularization parameter is selected from $\{0.1,0.01,0.001\}$ for online KFMC on all experiments.
Online KFMC also requires a momentum update, for which we take the author's suggested value $.5$.

\paragraph{Computational complexity}
For a new data vector in $\Rbb^p$ with $k$ observed entries, GROUSE has the smallest computation time $O(pr_g+kr_g^2)$ with rank $r_g<p$;
online EM comes second with computation time $O(k^3+k(p-k)p)$;
online KFMC has the largest computation time $O(r_k^3)$ with $r_k>p$.

\paragraph{Offline synthetic experiment}
We follow the setting in \cite{zhao2020missing}:
the $5$ continuous entries have exponential distribution with parameter $1/3$;
The cut points for generating the ordinal and binary entries are randomly selected.
The masking is done such that $2$ out of $5$ entries for each data type are masked. 
We generate independent two identical and independent datasets: one for choosing the tuning parameters, the other for training and evaluating the performance.
For GROUSE, the selected rank is 1 and the selected constant $c$ in decaying stepsize $c/t$ is 1.
For online KFMC, the selected rank is $400$ and the selected regularization is $.1$.
The used batch size is $100$ for all methods.

\paragraph{Online synthetic experiment} 
We also generate independent two identical and independent datasets: one for choosing the tuning parameters, the other for training and evaluating the performance.
For online EM, the selected window size is $200$.
For GROUSE, the selected rank is 1 and the selected constant stepsize is $10^{-6}$.
For online KFMC, the selected rank is $200$ and the selected regularization is $.1$.
The used batch size is $40$ for all methods.

\paragraph{Offline real data experiment} 
We divide all available entries into training ($80\%$), validation ($10\%$) and testing ($10\%$).
For GROUSE, the selected rank is 5 and the selected constant $c$ in decaying step size $c/t$ is 1. For online KFMC, the selected rank is $200$ and the selected regularization parameter is $.1$.
The used batch size is $121$ for all methods.

\paragraph{Online real data experiment}
Three stocks have missing entries ($90.6\%, 16.9\%$ and $35.6\%$) corresponding to dates before the stock was publicly traded.
We construct a dataset of size $5029\times 60$, where
the first $30$ columns store yesterday's price (or log return) and last $30$ columns store today's.
We scan through the rows, making online imputations.
Upon reaching the $t$-th row, 
the first $t-1$ rows and the first $30$ columns of the $t$-th row are completely revealed, while the last 30 columns of the $t$-th row are to be predicted and thus masked. 
Once the prediction is made and evaluated, the masked entries are revealed
to update the model parameters.
such dataset allows the imputations methods can learn 
both the dependence among different stocks and the auto-dependence of each stock.
We use first $400$ days' data to train the model and the next $400$ days' data as a validation set to choose the tuning parameters, and all remaining data to evaluate the model performance.
For online EM, the selected window size is $50$ for price prediction and $200$ for log return prediction.
For GROUSE, the rank and the constant step size are selected as $\{10,10^{-6}\}$ for price prediction, and $\{1,10^{-6}\}$ for log return prediction. 
For online KFMC, the rank and the regularization are selected as $\{300,0.1\}$ for both the price prediction and the log return prediction.
The used batch size is $40$ for all methods.

\section{Additional experiments}
\subsection{Acceleration of parallelism}
Here we report the acceleration achieved using parallelism for Gaussian copula methods.
We only report the runtime comparison on offline datasets, since the parallelism does not influence the algorithm accuracy.
We use 2 cores to implement the parallelism.
The results in \cref{table:runtime_parallel} show the parallelism brings considerable speedups for all Gaussian copula algorithms.

\begin{table}
\centering
\begin{adjustbox}{width=\columnwidth}
\begin{tabular}{ |l | c |c |}
\hline
 &  Offline Sim ($5000\times 15$)  & Movielens ($6027\times 207$)\\
\hline
Offline EM & 188(1), 87(2) & 1690(9), 781(2)\\
Minibatch EM  & 48(1), 28(0) & 252(2), 142(2)\\
Online EM & 52(0), 34(1)& 269(3), 169(11)\\
\hline
\end{tabular}
\label{table:runtime_parallel}
\end{adjustbox}
\caption{Mean(sd) runtime of Gaussian copula methods for offline datasets over $10$ trials. In each cell, the runtime with 2 cores follows that with 1 core.}
\end{table}

\subsection{Robustness to varying data dimension, missing ratio and missing mechanism}
We add experiments under MAR and MNAR and also experiments using missing ratios, number of samples, and variable dimensions in our online synthetic experiments. The original setting has $p=15$ variables, 6000 samples in total ($n=2000$ samples for each distribution period), and $40\%$ missing entries under MCAR. 
We vary each of these three setups: $n$, $p$, and missing ratio. We also design MNAR such that larger values have smaller missing probabilities, shown as in \cref{table:mnar_mechanism}.

The results in \cref{table:additional_sim} record the scaled MAE (SMAE) of the imputation estimator, as used in Table 1 in the manuscript. They show our method is actually robust to violated missing mechanisms and various number of samples, dimensions and missing ratios.
\begin{table*}[t]
\centering
%\begin{adjustbox}{width=\columnwidth}
\begin{tabular}{ |l | c |c |c|}
\hline
 Variable type&  20\% missing probability & 40\%  missing probability	& 60\% missing probability\\
\hline
Continuous  &  Entries above 75\% quantile & Entries between 75\% and 25\% quantiles & Entries below 25\% quantile\\
\hline
Ordinal & Entries equal to 5, 4	& Entries equal to 3 &	Entries equal to 2, 1\\
\hline
Binary &Entries equal to 1 & NA & Entries equal to 0\\
\hline
\end{tabular}
\label{table:mnar_mechanism}
%\end{adjustbox}
\caption{A MNAR mechanism used. For each variable, the missing probability of an entry solely depends on its own value. Entries with smaller values have high missing probabilities.}
\end{table*}

\begin{table*}[t]
\begin{center}
%\begin{adjustbox}{width=\textwidth}
%\begin{adjustbox}{width=2\columnwidth}
\begin{tabular}{|l|rrr|rrr|rrr|}
\hline
 & \multicolumn{3}{|c|}{20\% missing} & \multicolumn{3}{|c|}{40\% missing}  & \multicolumn{3}{|c|}{60\% missing} \\
\hline
Method &   Cont & Ord  & Bin &   Cont& Ord  & Bin &   Cont & Ord  & Bin \\ 
\hline
OnlineEM &  
\textbf{.76(.08)} & \textbf{.81(.10)} & \textbf{.63(.12)} & \textbf{.84(.07)} & \textbf{.89(.07)} & \textbf{.72(.11)} & \textbf{.91(.05)} & \textbf{.96(.04)} & \textbf{.83(.09)}\\
OnlineKFMC & 
.94(.06)&1.08(.38)&.79(.15)&.98(.06)&.17(.46)&.88(.12)&1.02(.07)&	1.32(.59)&.96(.10)\\
GROUSE &  
1.17(.06)&1.70(.36)&1.12(.11)&1.20(.07)&1.80(.40)&1.10(.06)&1.31(.14)&	2.09(.49)&1.12(.06) \\
\hline
 & \multicolumn{3}{|c|}{$n=1000$} & \multicolumn{3}{|c|}{$n=2000$}  & \multicolumn{3}{|c|}{$n=3000$} \\
\hline
Method &   Cont & Ord  & Bin &   Cont& Ord  & Bin &   Cont & Ord  & Bin \\ 
\hline
OnlineEM &  
\textbf{.87(.08)}&\textbf{.90(.09)}&\textbf{.75(.13)}&\textbf{.84(.07)}&\textbf{.89(.07)}&\textbf{.72(.11)}&\textbf{.83(.06)}&\textbf{.88(.07)}&\textbf{.70(.10)}\\
OnlineKFMC & 
1.01(.08)&1.29(.51)&.95(.13)&.98(.06)&1.17(.46)&.88(.12)&.97(.05)&1.10(.38)&.83(.13)\\
GROUSE &  
1.22(.09)&1.80(.45)&1.12(.06)&1.20(.07)&1.80(.40)&1.10(.06)&1.20(.06)&1.80(.34)&1.11(.06)\\
\hline
 & \multicolumn{3}{|c|}{$p=15$} & \multicolumn{3}{|c|}{$p=30$}  & \multicolumn{3}{|c|}{$p=45$} \\
\hline
Method &   Cont & Ord  & Bin &   Cont& Ord  & Bin &   Cont & Ord  & Bin \\ 
\hline
OnlineEM &  
\textbf{.84(.07)}&\textbf{.89(.07)}&\textbf{.72(.11)}&\textbf{.85(.07)}&\textbf{.90(.07)}&\textbf{.74(.10)}&\textbf{.86(.07)}&\textbf{.92(.08)}&\textbf{.76(.09)}\\
OnlineKFMC & 
.98(.06)&1.17(.46)&.88(.12)&.98(.05)&\textbf{.89(.11)}&.98(.06)&.98(.05)&.98(.05)&.98(.05)\\
GROUSE &  
1.20(.07)&1.80(.40)&1.10(.06)&1.13(.05)&1.12(.06)&1.14(.05)&1.13(.06)&1.13(.06)&1.12(.06)\\
\hline
 & \multicolumn{3}{|c|}{MCAR} & \multicolumn{3}{|c|}{MNAR}  & \multicolumn{3}{|c|}{}  \\
\hline
Method &   Cont & Ord  & Bin &   Cont& Ord  & Bin &   &&\\ 
\hline
OnlineEM & \textbf{.84(.07)}&\textbf{.89(.07)}&\textbf{.72(.11)}&\textbf{.89(.07)}&\textbf{.99(.07)}&.73(.05)&&\\
OnlineKFMC & 
.98(.06)&1.17(.46)&.88(.12)&.93(.05)&1.22(.54)&.70(.11)&&\\
GROUSE &  
1.20(.07)&1.80(.40)&1.10(.06)&1.46(.07)&1.95(.40)&\textbf{.66(.04)}&&\\
\hline
\end{tabular}
%\end{adjustbox}
\caption{Mean(sd) for runtime, imputation error for each data type for additional synthetic online data over $10$ trials. Figure 2 in the main paper corresponds to 40\% missing, $n=2000, p=15$ and MCAR here. }
\label{table:additional_sim}
%\end{adjustbox}
\end{center}
\end{table*}

\section{Proofs}
\subsection{Proof of Lemma 1}
\begin{proof}
First note 
\begin{align*}
    &Q(\Sigma;\Sigma^t, \{\xio\}_{t\in S_t})\\
    =&\frac{1}{|S_t|}\sum_{i\in S_t}\Ebb[\ell(\Sigma;\{\bz^i,\xio\}_{i\in S_t})|\xio, \Sigma^{t-1}, \hat\bigf]\\
    =&\frac{1}{|S_t|}\sum_{i\in S_t}\Ebb\left[c-\frac{\log|\Sigma|}{2}-\frac{(\bz^i)^\top\Sigma^{-1}\bz^i}{2}\Big|\xio, \Sigma^{t-1}, \hat\bigf\right]\\
    =&\Ebb\left[c-\frac{\log|\Sigma|
    +
    \tr\left(\Sigma^{-1}\frac{1}{|S_t|}\sum_{i\in S_t}\bz^i(\bz^i)^\top\right)}{2}\Bigg|\xio, \Sigma^{t-1}, \hat\bigf\right]\\
    =&c-\frac{\log|\Sigma|
    +
    \tr\left(\Sigma^{-1}
    \Ebb\left[\frac{1}{|S_t|}\sum_{i\in S_t}\bz^i(\bz^i)^\top|\xio, \Sigma^{t-1}, \hat\bigf\right]
    \right)}{2}\\
\end{align*}
where $c>0$ is a constant.

Denote 
\[
E_t=\Ebb\left[\frac{\sum_{i\in S_t}\bz^i(\bz^i)^\top}{|S_t|}|\xio, \Sigma^{t-1}, \hat\bigf\right]
\]
%$E_t:=\Ebb\left[\frac{\sum_{i\in S_t}\bz^i(\bz^i)^\top}{|S_t|}|\xio, \Sigma^{t-1}, \hat\bigf\right]$ as $E(\{\xio\}_{i\in S_t}, \Sigma^{t-1}, \hat\bigf)$.
It is easy to show through induction that $Q_t(\Sigma)$ can be written as 
\[
Q_t(\Sigma) = \sum_{l=1}^t\alpha^t_lQ(\Sigma;\Sigma^{l-1},\{\bx^i\}_{i\in S_l}), 
\]
with $\sum_{l=1}^t\alpha^t_l=1$ and $\alpha^t_l>0$.
Thus we have 
\[
Q_t(\Sigma) = c - \frac{\log|\Sigma|+\tr(\Sigma^{-1}\sum_{l=1}^t\alpha^t_lE_t)}{2}
\]
Then solving $\argmax_{\Sigma}Q_t(\Sigma)$ is the classical problem of the MLE of Gaussian covariance matrix, which yields $\Sigma^t = \argmax Q_t(\Sigma) = \sum_{l=1}^t\alpha^t_lE_l$, when $\sum_{l=1}^t\alpha^t_lE_l$ is positive definite.
Since we require $|S_l|>p$, we have $E_l$ as positive definite matrix and thus $\sum_{\ell=1}^t\alpha_l^tE_t$ is also positive definite.

For the first data batch, $\Sigma^1$ is estiamed as in the offline setting: 
$\Sigma^1 = E_1$ wit initial estimate $\Sigma^0$, thus we set $\gamma_0=1$ to satisfy $\Sigma^{t+1}=(1-\gamma_t)\Sigma^t+\gamma_t E_{t+1}$ for $t=0$.
For any $t>1$ and $\gamma_t\in (0,1)$,
note by the definition of $Q_t(\Sigma)$:
\[
\alpha_l^{t+1} = \alpha_l^t(1-\gamma_t), \mbox{ for } l=1,\ldots,t, \mbox{ and }\alpha_{t+1}^{t+1}=\gamma_t.
\]
then
\begin{align*}
    \Sigma^{t+1} &= \sum_{l=1}^{t+1}\alpha^{t+1}_lE_l=\sum_{l=1}^{t}\alpha^{t+1}_lE_l+\alpha^{t+1}_{t+1}E_{t+1} \\
    &= \sum_{l=1}^{t}\alpha_l^t(1-\gamma_t)E_l + \gamma_t E_{t+1} = (1-\gamma_t)\Sigma^t + \gamma_t E_{t+1}.
\end{align*}
which finishes the proof.
\end{proof}

\subsection{Proof of Theorem 1}
We first formally define some concepts, and then rigorously restate Theorem 1 with complete description.
In the proof, we show our Theorem 1 is a special case of Theorem 1 in \citet{cappe2009line} by verifying the their required assumptions  are satisfied in our setting.

\paragraph{Distribution function for mixed data}
For a mixed data vector $\bx = (\bx_{\indexC}, \bx_{\indexD})$ with $\bx_{\indexC}$ as continuous random variables and $\bx_{\indexD}$ as ordinal random variables,
we use the notion of distribution function for $\bx$ as $f(\bx)=f(\xc)\mathrm{P}(\xd)\in \Rbb$, with $f(\xc)$ as the PDF of $\xc$ and $\Prm(\xd)$ as the PMF (probability mass function) of $\xd$. 

\paragraph{Distribution over incomplete data}
Let $\tilde \bx=(\tilde x_1,...,\tilde x_p)=(\tilde \bx_{\indexO}, \tilde \bx_{\indexM})$ be a underlying complete vector that is observed at $\indexO\subset [p]$, $\bm$ be the associated observed-data indicator vector: $\bm=(m_1,..,m_p)$ where $m_j=1$ if $\tilde x_j$ is observed ($j\in \indexO$) and $m_j=0$ if $\tilde x_j$ is missing ($j\in \indexM$).
Also define $\bx=(x_1,...,x_p)$ be the incomplete version of $\tilde \bx$ with a special category \na $\mbox{ }$ at missing locations: $x_j=\tilde x_j$ if $m_j=1$ and $x_j=\na$ if $m_j=0$.
Denote the deterministic mapping from $(\tilde \bx, \bm)$ to $\bx$ as $T(\tilde \bx, \bm)=\bx$.

Once we are given an incomplete data vector $\bx$, the actual observation is $(\xobs, \bm)$.
Our goal is to learn the distribution associated with the underlying complete vector $\tilde \bx$ instead of the distribution of $\bx$, since the latter also requires characterizing the distribution of $\bm$.
Under the missing at random (MAR) assumption, we have 
\begin{align*}
    f(\xobs, \bm)&=\int f(\xobs, \xmis)\Prm(\bm|\xobs, \xmis)d\xmis\\
    &=\int f(\xobs, \xmis)d\xmis\Prm(\bm|\xobs)=f(\xobs)\Prm(\bm|\xobs).
\end{align*}

To distinguish a few definitions, there is a distribution $\pi^*(\tilde \bx)$ for the true underlying complete vector $\tilde \bx$, 
a (joint) distribution $\pi(\bx)$ over the observed entries $\xobs$ and the missing locations $\bm$,
and a (marginal) distribution $\pi(\xobs)$ over the observed entries $\xobs$.
There is a one-to-one correspondence between the complete data distribution $\pi^*(\tilde \bx)$ and the observed data distribution $\pi(\xobs)$, 
since $\pi(\xobs)$ is the marginal distribution of $\pi^*(\tilde \bx)$ over dimensions $\indexO$.
The joint distribution $\pi(\bx)=\pi(\xobs)\Prm(\bm|\xobs)$, further requires the conditional distribution of $\bm|\xobs$, which is unknown.
%The convergence result of our Theorem 1 is stated over the marginal distribution $\pi(\xobs)$ and $g_\Sigma(\xobs)$,
%but the proof will be over the joint distribution $\pi(\bx)$ and $g_\Sigma(\bx)$ by 

When we say true data distribution over the observed entries, we refer to the (marginal) distribution $\pi(\xobs)$.
With a Gaussian copula model $\gc$, we denote the underlying complete distribution as $g_\Sigma^*(\tilde \bx)$, and the distribution of observed data as $g_\Sigma(\xobs)$.
We further construct the 
joint distribution over $(\xobs,\bm)$ as $g_\Sigma(\bx)$, using the same conditional distribution $\Prm(\bm|\xobs)$ as in $\pi(\bx)$,
for the purpose of proof.
We ignore the dependence on $\bigf$ because it is kept fixed during EM iterations, while $\Sigma$ is updated at each iteration. 
Define $d(x,A) = \textup{inf}(y\in A, |x-y|)$ with $|\cdot|$ as the $\ell_2$ norm.
%Given data batches $, ..., \{\bx^i\}_{i\in S_t}$,
%define the online computed expectation $\hat s_{t+1} = (1-\gamma_t)\hat s_t$
Now we are ready to restate our Theorem 1.

\begin{theorem}
Let $\pi(\xobs)$ be the distribution function of the true data-generating distribution of the observations and $g_{\Sigma}(\xobs)$ be the distribution function of the observed data from $\textup{GC}(\Sigma, \bigf)$, assuming data is missing at random (MAR).
Let $\mathcal{L}=\{\Sigma\in S_{++}^p: \nabla_{\Sigma}\textup{KL}(\pi||g_\Sigma)=0\}$ be the set of stationary points of $\textup{KL}(\pi||g_\Sigma)$ for a fixed $\bigf$.
%Denote by $\mathcal{L}=\{\Sigma\in S_{++}^p: \nabla_{\Sigma}\textup{KL}(\pi||g_\Sigma)=0\}$.
Under the following conditions,
\begin{enumerate}[leftmargin=*]
\item $\bigf$ remains unchanged across EM iterations; for all continuous dimensions $j$,  the range of $f_j^{-1}$ is a subset of $[-C, C]$ for some $C>0$; for each ordinal dimension $j$, the step function $f_j$ only has finite number of steps, i.e. $f_j(z_j)$ has finite number of ordinal levels.
\item The step-sizes $\gamma_t\in (0,1)$ satisfy 
$\sum_{t=1}^\infty \gamma_t^2<\sum_{t=1}^\infty \gamma_t=\infty$.
    \item With probability $1$, $\lim \textup{sup} |\Sigma_t|<\infty$
and $\lim \textup{inf}\{d( \Sigma_t, ( S_{++}^p)^c)\}>0$.
\item Let $\bX_t$ (size $|S_t|\times p$) denote the data observation in the $t$-th batch with points i.i.d. $\pi(\bx)$, and $\bZ_t$ as the latent data matrix corresponding to $\bX_t$. For the set $\Gamma = \{s\in S_{++}^p: \Erm_\pi[\Erm_{\Sigma=s}[\bZ_t^\top\bZ_t|\bX_t]]=s\}$ and $w(\Sigma):=\textup{KL}(\pi||g_\Sigma)$, $w(\Gamma)$ is nowhere dense.
\end{enumerate}
%\[\lim_{t\xrightarrow[]{}\infty}\inf_{\Sigma \in \mathcal{L}}|\Sigma^{t} - \Sigma|=0 \mbox{ with probability }1, \quad \mbox{ for }\Sigma^t \mbox{ obtained in \cref{eq:online_update} with any fixed } \bigf.\]
then the iterates $\Sigma^{t}$
produced by our online EM (Algorithm 1 in the main paper) satisfies that $\lim_{t\xrightarrow[]{}\infty}d(\Sigma^t, \mathcal{L})=0$ with probability $1$.

% the distance from $\Sigma^{t}$, 
% obtained in \cref{eq:online_update}, to $\mathcal{L}$ converges to $0$ with probability $1$ as $t\xrightarrow{}\infty$.
\label{theorem:kl}
\end{theorem}

\begin{proof}
We first formally state our latent model: the observed variables and the latent variables, as well as the conditional distribution of the latent given the observed.
Then we show our stated convergence result is a special case of the result in Theorem 1 of \citet{cappe2009line}.
The following proof consists of showing our latent model satisfies the assumptions required in Theorem 1 of \citet{cappe2009line},
and thus our results hold according to Theorem 1 of \citet{cappe2009line}.

\paragraph{The employed latent model}
We treat our defined $\bx$ (with \na $\mbox{ } $at missing entries) as the ``observed variables'' in our latent model. 
Since $\bx = T(\tilde \bx, \bm)$ and there exists a Gaussian latent variable $\bz \sim \mathcal{N}(\bo, \Sigma)$ such that $\tilde \bx = \bigf(\bz)$,
we treat $(\bz, \bm)$ as our ``latent variables''.
Conditional on known $\bx$,  the distribution of $\bm$ reduces to a single point, denoted as $\bm_\bx$.
The distribution of $\bz$ is $\mathcal{N}(0, \Sigma)$ truncated to the region $\{\bz: T(\bigf(\bz), \bm_\bx)=\bx\}$ for $\tilde \bx$.
With slight abuse of notation, we write $\{\bz: T(\bigf(\bz), \bm_\bx)=\bx\}$ as $\bigf^{-1}(\bx)$ and $f_j^{-1}(\na)=\Rbb$ for any $j\in [p]$. Now note 
\begin{align*}
    \textup{KL}(\pi(\bx)||g_\Sigma(\bx)) &=\Erm_{\xobs, \bm}\log \frac{\pi(\xobs)\Prm(\bm|\xobs)}{g_\Sigma(\xobs)\Prm(\bm|\xobs)}\\
    &=\textup{KL}(\pi(\xobs)||g_\Sigma(\xobs)).
\end{align*}

Thus our stated result matches Theorem 1 of \citet{cappe2009line}: the online EM estimate converges to the set of stationary points of KL divergence between the true data distribution and the learned data distribution, on ``observed variables'' $\bx$.

\paragraph{Verification of Assumptions 1 in \citet{cappe2009line}}
For simplicity, assume each data batch has $n$ points. We drop the time index on the data points and simply write the data points in each batch as $\{\bx^i\}_{i=1,..,n}$ or the corresponding matrix $\bX$.
Also denote the latent data as $\{\bz^i, \bm^i\}_{i=1,..,n}$ and the corresponding matrix data $\bZ, \bM$.
The complete data likelihood for a batch is 
\begin{equation*}
    L(\Sigma;\{\bx^i, \bz^i, \bm^i\})=
    c
    e^{\log |\Sigma| + \frac{1}{n} \tr(\Sigma^{-1}\bZ^\top \bZ)},
\end{equation*}
where $c$ is a constant w.r.t. $\Sigma$:
\[
c=\prod_{i=1}^n1(T(\bigf(\bz^i), \bm^i)=\bx^i)f(\bm^i)
    (2\pi)^{-\frac{np}{2}}e^{-\frac{n}{2}}.
\]
Thus it belongs to the exponential family,
with sufficient statistics $s=\frac{1}{n}\bZ^\top \bZ$, $\psi(\Sigma) = -\log |\Sigma|$ and $\phi(\Sigma)=\Sigma^{-1}$.
Thus Assumption 1a is satisfied. 

Denote the sample space for $\Sigma$ as the set of all positive definite symmetric matrices: ${\Theta} = S_{++}^p$.
The function $\bar s(\bX;\Sigma):=\Erm_{\Sigma}[\bZ^\top \bZ|\bX]$ is well defined for all $\bX$ and all $\Sigma \in \Theta$.
To see why, note $\Erm[z_j|x_j]$ and $\Varrm[z_j|x_j]$ have finite closed form expression for any $x_j$,
and thus $\Erm[z_j^2|x_j]$ has finite closed form expression for any $x_j$ for all $j\in[p]$.
Then for any $i,j\in [p]$, $\Erm[z_iz_j|\bx]$ is finite and thus well defined using Cauchy inequality.
Thus Assumption 1b is satisfied. 

Let $\mathcal{S}=S_{++}^p$, then clearly $\mathcal{S}$ is a convex open subset of all symmetric matrices.
For any $\Sigma, \Sigma' \in \mathcal{S}$, any $\bX$ and $\gamma \in [0,1)$,
we have $\bar s(\bar X; \Sigma)$ as positive semidefinite matrix and thus $(1-\gamma)\Sigma'+\gamma \bar s(\bX;\Sigma) \in \Scal$.
In our situation, $\ell(s;\Sigma)=\log |\Sigma|+\tr(\Sigma^{-1}s)$. 
Note solving $\max_{\Sigma \in \Scal}\ell(s;\Sigma)$ is equivalent to solving the MLE of multivariate normal covariance.
By classical results, 
for any $s\in \Scal$, $\ell(s;\Sigma)$ has a unique global maximum over $\Theta$ at $\Sigma = s$, denoted as $\bar \Sigma(s) = s$.
Thus Assumption 1c is satisfied. 

\paragraph{Verification of Assumptions 2 in \citet{cappe2009line}}
For (2a), in our situation, $\psi(\Sigma) = -\log |\Sigma|$ and $\phi(\Sigma)=\Sigma^{-1}$ are clearly twice continuous differentiable in $\Theta = S_{++}^p$.
For (2b), $\bar \Sigma(s)$ is simply the identity function and thus continuously differentiable in $\Scal$.
For (2c), first note 
\[
\bar s(\bX; \bar \Sigma(s)) = \bar s(\bX; s) = \Erm_{\Sigma = s}[\bZ^\top \bZ|\bX]
\]
Now we bound the max-norm of $\Erm_{\Sigma = s}[\bZ^\top \bZ|\bX]$ uniformly over all possible $\bX$ for a given $s$.
To do so, it suffices to bound the max-norm of $\Erm_{\Sigma = s}[\bz^i (\bz^i)^\top|\bx^i]$ for a single point $i$.
We ignore $i$ for notation simplicity.
It suffices to bound the max-norm of the diagonal entries, since we can bound the max-norm all off-diagonal entries using the max-norm of the diagonal entries through Cauchy-Schwarz inequality.
Now if $x_j$ is an observed continuous entry, by regularity condition on $\bigf$, we have $z_j = f_j^{-1}(x_j) \in [-C, C]$ and thus $\Erm_{\Sigma=s}[z_j^2|x_j]$ is finite.
If $x_j$ is a missing entry, we have $\Erm_{\Sigma=s}[z_j^2|x_j]=s_{jj}^2$, the $(j,j)$-th entry of $s$.
At last if $x_j$ is an observed ordinal entry, we have 
\begin{align*}
    \Erm_{\Sigma=s}[z_j^2|x_j] =&\frac{\int_{z_j \in f_j^{-1}(x_j)}z_j^2\phi(z_j;0,s_{jj}^2)dz_j}
{\int_{z_j \in f_j^{-1}(x_j)}\phi(z_j;0,s_{jj}^2)dz_j}\\
\leq & C_j\int_{z_j \in f_j^{-1}(x_j)}z_j^2\phi(z_j;0,s_{jj}^2)dz_j\\
\leq & C_j\int_{z_j \in \Rbb}z_j^2\phi(z_j;0,s_{jj}^2)dz_j =C_js_{jj}^2.
\end{align*}
where $C_j=\frac{1}{\min_j\int_{z_j \in f_j^{-1}(x_j)}\phi(z_j;0,s_{jj}^2)dz_j}$.
Note $C_j$ is finite and depends only on the step-wise function $f_j$ which has only finite number of steps.
Thus we can uniformly bound the max-norm of diagonal entries of $\Erm_{\Sigma=s}[z_j^2|x_j]$ using diagonal entries of $s$.
For all compact subsets $\mathcal{K}\subset \Scal$, and for all $s\in \mathcal{K}$, the diagonal entries of $s$ are bounded and thus $\Erm_{\Sigma = s}[\bZ^\top \bZ|\bX]$ are bounded.
In other words, we can bound $\sup_{s\in \mathcal{K}}|\bar s(\bX; \bar \Sigma(s))|$ uniformly over $\bX$ for a given $\mathcal{K}$.
Using similar arguments, for any $k>2$, we can bound $\sup_{s\in \mathcal{K}}|\bar s(\bX; \bar \Sigma(s))|^k$ uniformly over $\bX$ for a given $\mathcal{K}$
and thus $\Erm_\pi(\sup_{s\in \mathcal{K}}|\bar s(\bX; \bar \Sigma(s))|^k)<\infty$ for fixed $\bigf$.

%We show $\lim \textup{sup}|\hat s_n|<\infty$ by induction. 
%Suppose the max-norm of the diagonal entries in $\hat s_n$ is bounded by a sufficiently large constant $C$.
%Then
%\begin{align*}   |\hat s_{n+1}| &= |(1-\gamma_{n+1})\hat s_n + \gamma_{n+1}\bar s(\bx; \Sigma_n)| \leq (1-\gamma_{n+1})C + \gamma_{n+1}|\bar s(\bx; \hat s_n)| \\    & \leq (1-\gamma_{n+1})C + \gamma_{n+1}C' C 
%\end{align*}
Now we have verified all the assumptions of Theorem 1 in \citet{cappe2009line} which we do not include as our assumptions,
and thus finishes the proof. 
\end{proof}
\paragraph{Discussion on our assumptions}
Our assumption 1 is easily satisfied in practice. Once with access to moderate number of data points, once can pre-compute $\bigf$ and fix it among EM iterations. Our experiments show $200$ data points can provide good performance.
In practice, one use the scaled empirical CDF  to estimate $f^{-1}$, which ensures $f_j^{-1}$ to be a subset of $[-C, C]$ for sufficiently large $C$ (depending on data size $n$).
Also it is reasonable to model all ordinal variables to have finite number of ordinal levels, since we can only observed finite number of levels in practice.

Our assumptions 2-4 follow the assumptions in Theorem 1 of \citet{cappe2009line}.
Assumption 2 is standard for decreasing step size stochastic approximation and $\gamma_t=c/t$ with some constant $c>0$ satisfies the condition.
Assumption 3 corresponds to a stability assumption which is not trivial.
In practice, we enforce the stability by projecting the estimated covariance matrix to a correlation matrix.

%\bibliography{aaai22}

\end{appendix}
\end{document}